%% file: 0_main.tex
\documentclass{article}

\usepackage[preprint]{tmlr}

\usepackage{floatrow}

\usepackage[utf8]{inputenc} %
\usepackage[T1]{fontenc}    %
\usepackage{url}            %
\usepackage{booktabs}       %
\usepackage{amsfonts}       %
\usepackage{nicefrac}       %
\usepackage{microtype}      %

\input{math_commands.tex}

\usepackage{sidecap}

\sidecaptionvpos{figure}{c}

\usepackage{url}

\usepackage{algorithm}
\usepackage{algpseudocode}

\include{2_draftSwitch}

\include{_utility}

\newcommand{\githubRepo}{\url{https://github.com/SamD770/Generative-Models-Knowledge}}

\newcommand{\papertitle}{Approximations to the Fisher Information Metric of\\ Deep Generative Models for Out-Of-Distribution Detection}
\title{\papertitle}

\author{Sam Dauncey $^{1}$, Chris Holmes $^{2,3}$, Christopher Williams $^2$ \& Fabian Falck $^{2,3}$ \\
$^1$University of Edinburgh, $^2$University of Oxford, $^3$The Alan Turing Institute  \\
\texttt{s.dauncey@sms.ed.ac.uk},
\texttt{\{cholmes, williams, fabian.falck\}@stats.ox.ac.uk} 
}

\newcommand{\scorevec}[2]{\nabla_{#2} l(#1)}

\newcommand{\scorevectheta}[1]{\scorevec{#1}{\B{\theta}}}
\newcommand{\scorevecthetalayer}[1]{\scorevec{#1}{\B{\theta}_j}}

\newcommand{\modeldist}{p^{\B{\theta}}}

\begin{document}

\maketitle

\begin{abstract}

Likelihood-based deep generative models such as score-based diffusion models and variational autoencoders are state-of-the-art machine learning models approximating high-dimensional distributions of data such as images, text, or audio.  %
One of many downstream tasks they can be naturally applied to is out-of-distribution (OOD) detection. 
However, seminal work by Nalisnick et al. which we reproduce showed that deep generative models consistently infer higher log-likelihoods for OOD data than data they were trained on, marking an open problem.
In this work, we analyse using the gradient of a data point with respect to the parameters of the deep generative model for OOD detection, based on the simple intuition that OOD data should have larger gradient norms than training data.
We formalise measuring the size of the gradient as approximating the Fisher information metric. 
We show that the Fisher information matrix (FIM) has large absolute diagonal values, motivating the use of chi-square distributed, layer-wise gradient norms as features. We combine these features to make a simple, model-agnostic and hyperparameter-free method for OOD detection which estimates the joint density of the layer-wise gradient norms for a given data point.
We find that these layer-wise gradient norms are weakly correlated, rendering their combined usage informative, and prove that the layer-wise gradient norms satisfy the principle of (data representation) invariance.  %
Our empirical results indicate that this method outperforms the Typicality test for most deep generative models and image dataset pairings.

\end{abstract}

\section{Introduction}
\label{sec:Introduction}

Neural networks can be highly confident but incorrect when given inputs different to the distribution of data they were trained on \citep{szegedy2014intriguing, nguyen2015deep}.
While domain generalisation \cite{zhou2021domain} and domain adaptation \citep{garg2023rlsbench,DANN} methods tackle this problem by learning machine learning systems which are robust or can actively adapt to domain shift, there may be scenarios where for specific data points this domain shift is too severe to draw reliable inferences.
Identifying and possibly filtering out anomalies or \textit{out-of-distribution (OOD)} inputs before deploying the model in the wild is a viable strategy in such cases, especially for safety-critical applications \citep{ulmer2020trust, stilgoe2020uber, baur2021autoencoders}.

\textit{Deep generative models} such as variational autoencoders \cite{kingma2014vae}, normalising flows \cite{papamakarios2021normalizing}, autoregressive models \cite{van2016conditional,salimans2017pixelcnn++} and diffusion models \cite{sohl2015deep,ho2020denoising} are an important family of models in machine learning which allow us to generate high-quality samples from high-dimensional, multi-modal conditional or unconditional distributions in domains such as images, videos, text and speech.
Many current state-of-the-art methods are probabilistic: they approximate the data log-likelihood, the likelihood of a data sample given the learned model parameters under the data distribution, or a lower-bound thereof.
This renders them a natural candidate for the task of OOD detection as they `out-of-the-box' provide an OOD metric \cite{bishop1994novelty} (i.e. the approximated data likelihood) which they use the training objective.
However, \citet{nalisnick2018deep, choi2018waic} showed that many of the above mentioned classes of probabilistic deep generative models consistently infer \textit{higher} log-likelihoods for data points drawn from OOD datasets (non-training data) than for in-distribution (training data) samples. 
In Fig. \ref{likelihoodHistogram} we replicated their results with a Glow model, and show that score-based diffusion models are likewise affected by the phenomenon.  %
This result is very surprising given that generative models are trained to maximise the log-likelihood of the training data, and are able to generate high-fidelity, diverse samples from the training distribution.
This marks an open problem for deep generative models, and renders the direct use of their estimated likelihood in out-of-distribution detection infeasible. 
It also questions what deep generative models learn during training, and how they generalise.  

\begin{SCfigure}
    \centering
    \includegraphics[width=0.4\textwidth]{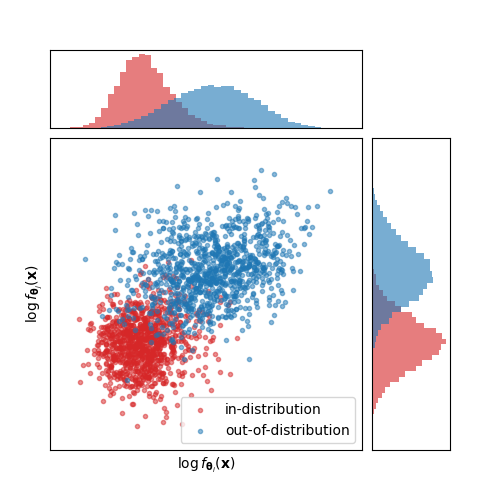}
    \caption{\textit{Gradients from certain layers are highly informative for OOD detection}. 
    We select two highly informative neural network layers of a deep generative model with parameters $\B{\theta}_i, \B{\theta}_j$ from a Glow \cite{kingma2018glow} model trained on \texttt{CIFAR-10} and plot the $L^2$-norm of the gradients $f_{\B{\theta}_j} = \norm{\nabla_{\B{\theta_{j}}}l(\B{x})}$ for in-distribution (\texttt{CIFAR-10}) and out-of-distribution (\texttt{SVHN}) samples $\B{x}$.
    The gradient norm of these two layers allows to separate in- and out-of-distribution samples. 
    }
    \label{fig:teaser}
\end{SCfigure}

Related work tackled this problem from two angles: 
explaining why log-likelihood estimates of these models fail to discriminate,  \cite{kirichenko2020flows, zhang2021understanding, lan2021, caterini2022entropy}, and   %
proposing likelihood-based OOD detection methods and adaptations of existing ones which may overcome these shortcomings \cite{ren2019, choi2018waic, hendrycks2019,  liu2020energy, havtorn2021, nalisnick2019}.
In \S \ref{sec:Related work}) we will analyse the small amount of previous work on gradient-based OOD detection, linking seemingly independent discoveries of other authors.

\textbf{Motivation and Intuition. }
This paper presents an alternative approach for OOD detection which we motivate in the following.
Consider the example of a (linear) regression model fitted on some (in-distribution) training data. 
If we now include an outlier in our training data and refit the model, the outlier will have a lot of influence on our estimate of the model's parameters compared to other in-distribution data points.
One way to formalise this intuition is using the \textit{hat-value}: 
The \textit{hat-value} is defined as the derivative $\frac{d \hat{y}}{d y}$ of the model's prediction $\hat{y}$ with respect to a given (dependent) data point $y$. 
It describes the leverage of a single data point, which is used for fitting the model, on the model's prediction for that data point after fitting.

We extend this intuition of OOD to deep learning by considering the gradient of the log-likelihood of a deep generative model with respect to its parameters, also known as the \textit{score}.   %
If a neural network converged to a local (or global) minimum, we expect the gradient of the likelihood with respect to the model's parameters to be flat for training data points.
Hence, the score is small in norm, and performing an optimiser step for a training data point would not change its parameters much.
If after many epochs of training we were now to present the model an OOD data point which it has not seen before, we expect the gradient---resembling the `hat-value of a neural network'---to be steep: the norm of the score would be large, just like hat values are large (in absolute value) for OOD data. 
An optimizer step with an OOD data point would change the neural network's parameters a lot.
It is this intuition which motivates us to theoretically analyse the use of the gradient for OOD detection.
In preview of our analyses, in Fig. \ref{fig:teaser} (and later more thoroughly in Fig. \ref{gradientHistograms}) we will precisely observe that (layer-wise) gradient norms are in general larger for OOD than for in-distribution data, which enables the use of gradients for OOD detection.
Code to reproduce our experimental results is publicly available on GitHub [anonymised during submission] \footnote{\githubRepo}.

Our contributions are as follows: 
(a) We analyse the use of the gradient of a data point with respect to the parameters of a deep generative model for OOD detection, and formalise this as approximating the Fisher information metric, a natural way of measuring the size of the gradient.
(b) We show that the Fisher information matrix (FIM) has large absolute diagonal values, motivating the use of layer-wise gradient norms which are chi-square distributed as a possible approximation.
Our theoretical results show that layer-wise gradients satisfy the principle of (data representation) invariance \cite{lan2021}, a desirable property for OOD methods.
We also find that these layer-wise gradient norms are weakly correlated, making their combined usage more informative.
(c) We propose a first simple, model-agnostic and hyperparameter-free method which estimates the joint density of layer-wise gradient norms for a given data point.
In our experiments, we find that this method outperforms the Typicality test for most deep generative models and image dataset pairings.

\section{Current Methods for OOD detection}

In this section, we define the OOD detection problem, describe the open problem of using deep generative models for OOD detection and how the input representation may explain this, and how a gradient-based method can be a compelling approach which is invariant to the input representation.

\subsection{OOD detection: Problem Formulation}
\label{sec:problem formulation}

Given training data $\B{x}_1 \dots \B{x}_N$ drawn from a distribution $p$ over the input space $\mathcal{X} \subseteq \mathbb{R}^D$, we define the problem of OOD detection as assigning an OOD score $S(\B{x})$ to each $\B{x} \in \mathcal{X}$ such that points with low OOD scores are semantically similar to points sampled from $p$. 
OOD detection is \emph{unsupervised} if it is not given class label information at training time.

The specific problem we are interested in is leveraging recent advances in deep generative models for unsupervised OOD detection. 
Here a deep generative model $p^{\B{\theta}}$ is trained to approximate the distribution of some training data $\B{x}_1 \dots \B{x}_N \sim p$, and $S$ is a statistic derived from $p^{\B{\theta}}$ (such as the model likelihood \cite{nalisnick2019}, a latent variable hierarchy \cite{Schirrmeister2020heirachy, havtorn2021}, or combinations thereof \cite{pmlr-v130-morningstar21a}).

In order to evaluate an OOD detection method, one is required to select semantically dissimilar surrogate out-distributions (e.g. a different dataset) to test against. 
Previous work has sought to define OOD detection as a generic test against data sampled from any differing distribution \cite{hendrycks2017a}. 
Our additional requirement that the out-distribution is semantically dissimilar is motivated by recent theoretical work by \cite{zhang2021understanding} showing that a single-sample test against all out-distributions is impossible.

\subsection{Likelihood-based methodology for unsupervised OOD detection}
\label{sec:Related work}

\begin{figure}[t]
    \centering
    \includegraphics[width=.35\textwidth]{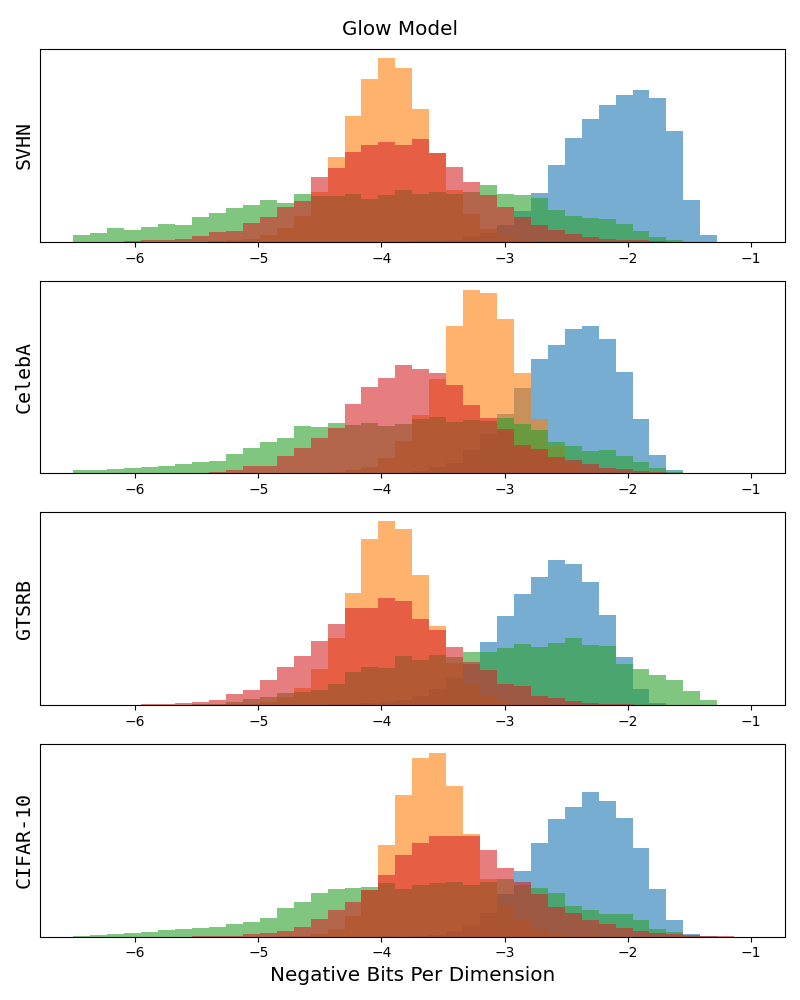}
    \includegraphics[width=.35\textwidth]{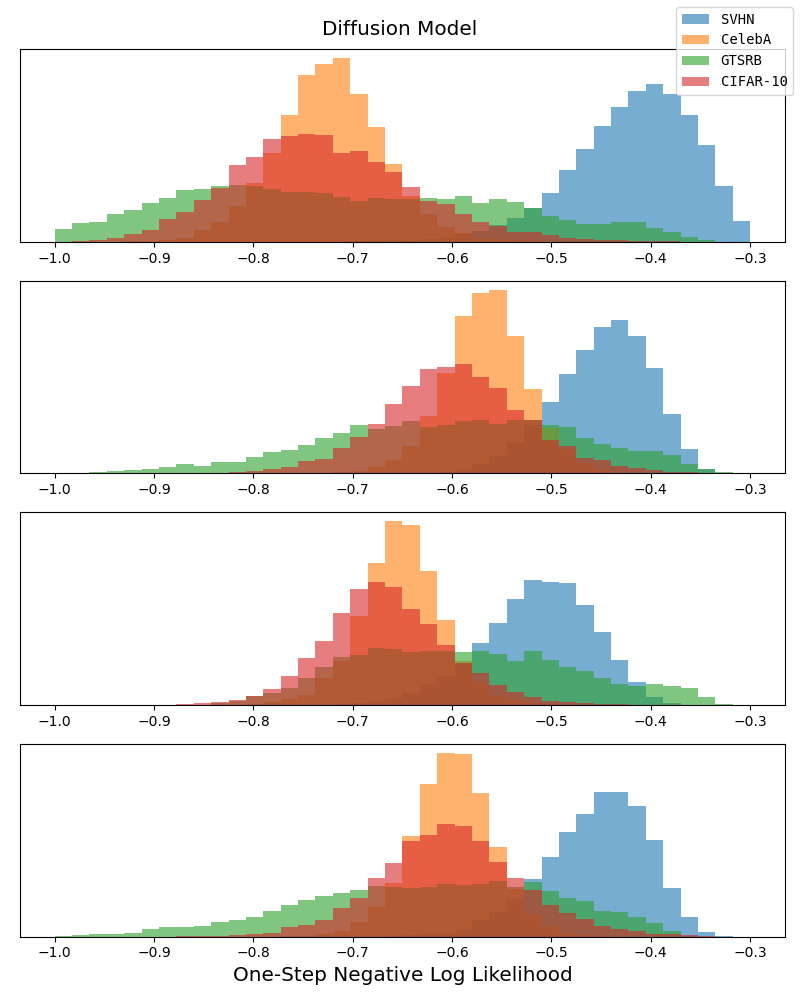}
    \caption{\emph{Counter-intuitive properties of likelihood-based generative models.} 
    Histogram of the negative log-likelihoods inferred from a Diffusion \cite{ho2020denoising} model [Left] and a Glow \cite{kingma2018glow} model [right] trained on one of four image datasets (corresponding to the four subplots) and evaluated on the test set of all four datasets, respectively. For diffusion models we use the negative log-likelihood from one step of the diffusion process $p^{\B{\theta}}(\B{x_0} \vert \B{x}_1)$. For both models we scale the log-likelihoods by the dimensionality of the data, in this case $3 \times 32 \times 32$.
    This Figure replicates the results in the seminal paper by \citet{nalisnick2018deep}, noting that our results for diffusion models are novel.
    We find that the training dataset has a counter-intuitively small impact on the ordering of the datasets as ranked by log-likelihood.
    \label{likelihoodHistogram}}
\end{figure}

\paragraph{Likelihood thresholding.} \cite{bishop1994novelty} proposed using the negative leaned model likelihood as an OOD score $S(\B{x}) = - \log p^{\B{\theta}}(\B{x})$.  
In their seminal paper, Nalisnick et al. empirically demonstrated that this approach fails for a wide variety of deep generative models \citep{nalisnick2018deep}. 
In particular they showed that certain image datasets such as \texttt{SVHN} are assigned systemically higher likelihoods than other image datasets such as \texttt{CIFAR10}, independent of the training distribution. 
We replicate this result (for a Glow model, a type of normalising flow, and for the first time a denoising diffusion model) in Figure \ref{likelihoodHistogram}. 
In their follow up work Nalisnick et al. argue from the Gaussian annulus theorem \cite{gaussianAnnulus} that samples with likelihoods much higher than likelihoods of in-distribution samples must be semantically atypical \citep{nalisnick2019}.
They use this to motivate OOD scoring based on the likelihood being too high or too low, defining the typicality \cite{typicalSetDefinition} score as $S(\B{x}) = \vert \log p^{\B{\theta}}(\B{x}) - \hat{\mathbb{H}} \vert $ , where $\hat{\mathbb{H}}$ is the average log-likelihood on some held-out training data.

\paragraph{Likelihood ratios. } 
The likelihood assigned by deep generative models has been shown to strongly correlate with complexity metrics such as the compression ratio achieved by simple image compression algorithms \cite{serra2019input}, and likelihoods from other generative models trained on highly diverse image distributions \cite{Schirrmeister2020heirachy}, with the highest likelihoods being assigned to constant images. 
To add to these findings, in Appendix \ref{proof:volumeTotalVariation} we use a very simple complexity metric $TV$, the total variation achieved by considering the image as a vector in $[0, 1]^{784}$, to show that the whole of \texttt{MNIST} is contained in a set of bounded complexity with volume (Lebesgue measure) $10^{-116}$. 
Thus a model needs to only assign a very low prior probability mass to this set for high likelihoods to be achieved, demonstrating the important connection between volume, complexity and model likelihoods which we hence discuss in \S \ref{sec: representation dependence}.
\cite{ren2019} argue in favour of using likelihood ratio tests in order to factor out the influence of the ``background likelihood'', the model's bias towards assigning high likelihoods to images with low complexity. 
In practice, this requires modelling the background likelihood via corruption of training data \cite{ren2019}, out-of-the-box and neural compressors \cite{serra2019input, zhang2021nelloc} or the levels of a model's latent variable heirarchy \cite{Schirrmeister2020heirachy, havtorn2021}, leading to restrictions for the data modalities or models to which the method can be applied to. 

\subsection{Representation dependence of the likelihood}

\label{sec: representation dependence}

\cite{lan2021} emphasise that the definition of likelihood requires choosing a method of assigning volumes to the input space $\mathcal{X}$. Specifically, datapoints could be represented as belonging to some other input space $\mathcal{T}$, linked via a smooth invertible coordinate transformation $T: \mathcal{X} \rightarrow \mathcal{T}$. 
The model probability density for a given datapoint $\B{x} \in \mathcal{X}$, which we denote $\modeldist_{\mathcal{X}}(\B{x})$, will thus differ from the probability density $\modeldist_{\mathcal{T}}(\B{t})$ of the corresponding point $\B{t} = T(\B{x})$ by a factor of the Jacobian determinant of $T$ \citep{lan2021}:
\begin{align}
    \modeldist_{\mathcal{T}}(\B{t}) = \modeldist_{\mathcal{X}}(\B{x}) 
    \; \left\lvert 
    \frac{\partial T }{\partial \B{x}} 
    \right\rvert^{-1}. \label{eq:change-of-vars}
\end{align}
\begin{figure}[t]
\floatbox[{\capbeside\thisfloatsetup{capbesideposition={right,top},capbesidewidth=0.57\textwidth}}]{figure}[\FBwidth]
{\caption{\textit{The log-likelihood heavily depends on data representation} \cite{lan2021}. Here we plot the first two samples of the \texttt{CIFAR10} dataset and the difference in Bits Per Dimension (BPD) induced by changing from an RGB to an HSV colour model: 
$\Delta^{RGB \to HSV}_{BPD} = \frac{\log_2 p_{RGB}(\mathbf{x}) - \log_2 p_{HSV}(\mathbf{x})} {3 \times 32 \times 32}.$
In Appendix \ref{app:Additional RGB-HSV}, we provide experimental details and inFig. \ref{fig:RGB-HSV appendix} replicate this for the first 20 samples, where we observe ${\Delta^{RGB \to HSV}_{BPD}}$ values ranging from $0.18$ to $1.76$
}
\label{fig:RGB-HSV}}
{\includegraphics[width=.4\textwidth]{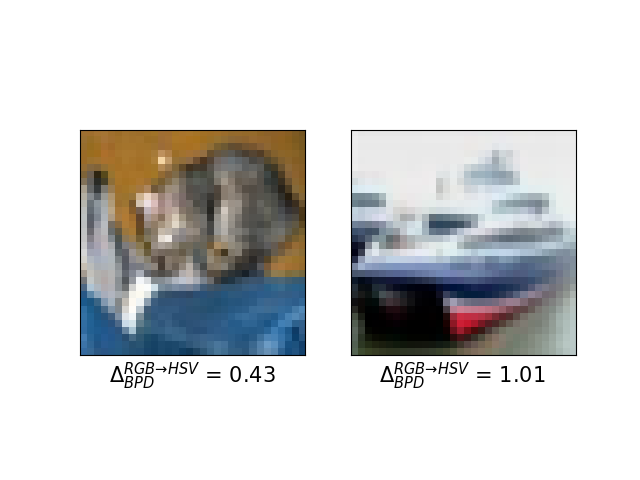}}
\end{figure}
The \emph{volume element} $\left\lvert \frac{\partial T }{\partial \B{x}} \right\rvert^{-1}$ describes the change of volumes local to $\B{x}$ as it is passed through $T$. 
This term can grow or shrink exponentially with the dimensionality of the problem, making its effect counter-intuitively large. 
As an empirical example in the case of image distributions, in Fig. \ref{fig:RGB-HSV} and Appendix \ref{app:Additional RGB-HSV} Fig \ref{fig:RGB-HSV appendix} we consider the case of a change of color model $T^{RGB \to HSV}$ from a Red-Green-Blue (RGB) to Hue-Saturation-Value (HSV) representation. 
We compute the induced change in bits per dimension as a scaled log-value of the volume element $\Delta^{RGB \to HSV}_{BPD} = \frac{1}{3 \times 32 \times 32} 
 \log \left\lvert \frac{\partial T^{RGB \to HSV} }{\partial \B{x}} \right\rvert$ and report values for $20$ non-cherry picked \texttt{CIFAR}-10 images ranging from $0.18$ to $1.76$. 
For comparison the average BPD reported in the seminal paper by \cite{nalisnick2018deep} was $3.46$ on \texttt{CIFAR10}, compared to $2.39$ on \texttt{SVHN} when evaluating with the same model.
Hence, if we use the likelihood for OOD detection, whether we classify a sample as OOD or not may flip for some samples merely by changing how the data is represented.

Motivated by the strong impact of the volume element, \cite{lan2021} propose a \textit{principle of (representation) invariance}: 
given a perfect model $p^{\B{\theta}^*}$ of the data distribution, the outcome of an unsupervised OOD detection method should not depend on how we represent the input space $\mathcal{X}$. In theory likelihood ratios are representation-invariant \citep{lan2021}, however in practice the method used to generate the background distribution often re-introduces dependence on the representation. 
For example \cite{ren2019} propose to generate the background distribution by re-sampling randomly chosen pixels as independent uniform distributions, re-introducing the notion of volume.

\subsection{Invariance of the gradient under invertible transformations}

\label{sec:scoreVecInvar}

To achieve a representation invariant OOD score \citep{lan2021}, we are thus motivated to quotient out the effect of the volume element in Eq. \eqref{eq:change-of-vars}. 
We now present our first theoretical contribution, which shows that methods based on the gradient of the log-likelihood do precisely this.

\begin{prop}
    Let $\modeldist_{\mathcal{X}}(\B{x})$ and $\modeldist_{\mathcal{T}}(\B{t})$ be two probability density functions corresponding to the same model distribution $\modeldist$ being represented on two different measure spaces $\mathcal{X}$ and $\mathcal{T}$. Suppose these representations encode the same information, i.e. there exists a smooth, invertible reparameterization $T: \mathcal{X} \rightarrow \mathcal{T}$ such that for $\B{x} \in \mathcal{X}$ and $\B{t} \in \mathcal{T}$ representing the same point we have $T(\B{x}) = \B{t}$.
    Then, the gradient vector $\nabla_{\B{\theta}} ( \log \modeldist)$ is invariant to the choice of representation, and in particular, 
    $\nabla_{\B{\theta}} ( \log \modeldist_{\mathcal{T}} )( \B{t}) 
    =
    \nabla_{\B{\theta}}  (\log \modeldist_{\mathcal{X}} ) (\B{x})$.
    \label{prop:gradient_invariance}
\end{prop}

\textbf{Proof.} See Appendix \ref{sec: proof_gradient_invariance}.
We prove analagous results for variational lower bounds (e.g. the ELBO of a VAE) in Appendix \ref{elboInvar}.

\begin{remark}
    Training a generative model $p^{\B{\theta}_0}$ with initialisation parameters ${\B{\theta}_0}$ with log-likelihood as the loss via gradient-descent produces training trajectories $\B{\theta}_0, \B{\theta}_1, \dots \B{\theta}_N$ which are representation-invariant.
    \label{remark:SGD-invariance}
\end{remark}

The interpretation of the above results is subtle.
We would like to caution the reader by noting it does \emph{not} mean that the inductive biases are discarded when the gradient is computed as inductive biases pertaining to distances between data points are frequently encoded in the parameter space. 
Further, remark \ref{remark:SGD-invariance} may explain why the likelihood can still be used to train deep generative models and allow them to generate convincing samples when using a gradient-based optimisation algorithm, even though the likelihood value itself appears uninformative for detecting if data is in-distribution.

\section{Methodology}

\label{sec:Methodology}

In this section, we develop a mathematically-principled method for gradient-based OOD detection.

\subsection{Layer-wise gradients are highly informative and differ in size by orders of magnitudes}

\label{sec:Layer-wise gradients are highly informative and differ in size by orders of magnitudes}

\begin{figure}[t]
    \centering
    \includegraphics[width=.9\textwidth]{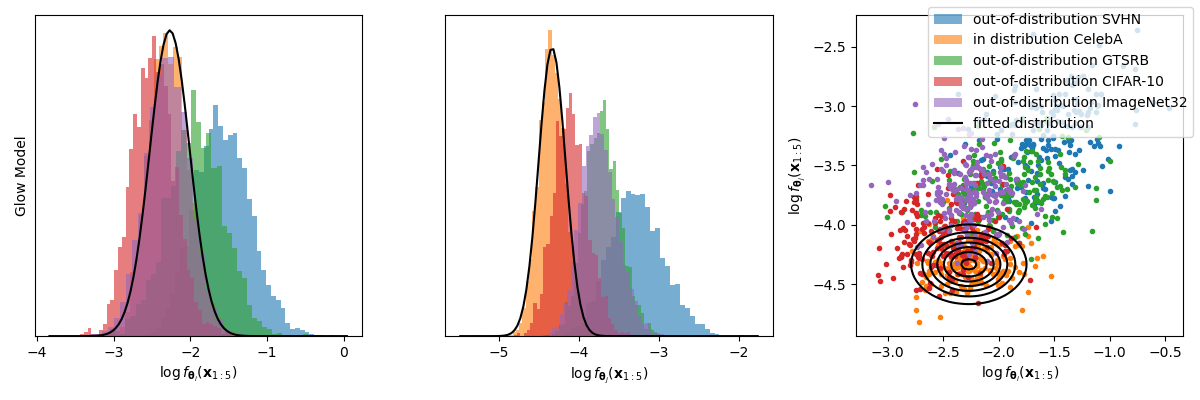}
    \includegraphics[width=.9\textwidth]{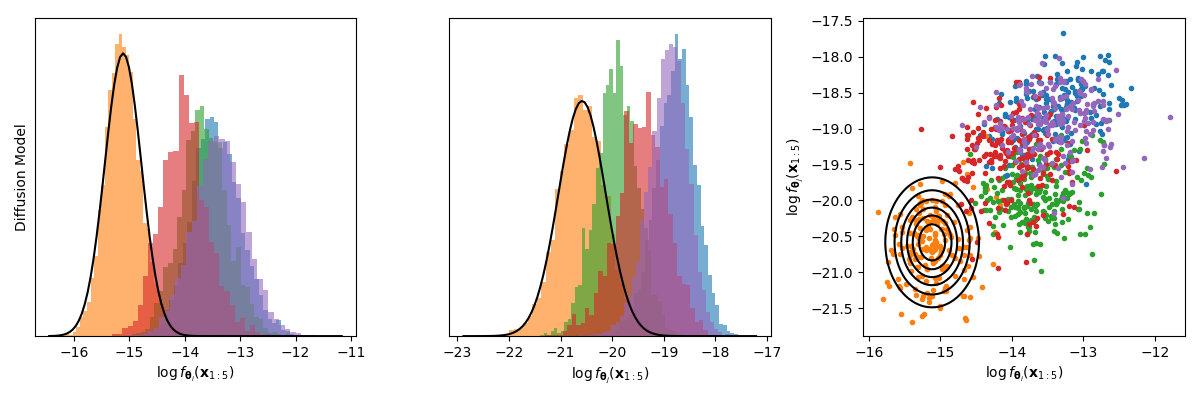}
    \caption{
    \textit{Layer-wise gradients of the log-likelihood (the score) are highly informative for OOD detection.}  
    Their size differs by orders of magnitudes between layers, and they are not strictly correlated, rendering layer-wise gradients (in contrast to the full gradient) discriminatory features for OOD detection.
    In each row, we randomly select two layers with parameters $\B{\theta}_i$, $\B{\theta}_j$ from a Glow \cite{kingma2018glow} model [Top] or a Diffusion model \cite{ho2020denoising} [Bottom], which have $1353$ and $276$ layers, respectively. 
    The models are trained on \texttt{CelebA}, a dataset that has proved challenging for OOD detection in previous work \cite{nalisnick2019}.
    We then evaluate this model on batches $\B{x}_{1:B}$ ($B=5$) drawn from the in-distribution and OOD test datasets and compute the squared layer-wise $L^2$-norm of the gradients of the log-likelihood with respect to the parameters of the layer, i.e. $f_{\B{\theta}_{j}}(\B{x}_{1:B}) = \norm{
    \nabla_{\B{\theta}_j} (\sum_{b=1}^B l(\B{x}_b))}_2^2$.  %
    [Left and Middle] shows the two layer-wise gradients separately, [Right] shows their interaction in a scatter plot.
    In Appendix \ref{app:Additional experimental details and results} Figures \ref{fig: gradientHistograms_app1} - \ref{fig: gradientHistograms_app3}, we provide our complete results, showing more layers from three likelihood-based generative models, each trained and evaluated on five datasets.}
    
    \label{gradientHistograms}
\end{figure}

We are now interested in formulating a method which uses the intuitively plausible (see \S \ref{sec:Introduction}) and data representation-invariant (see \S \ref{sec:scoreVecInvar}) score $\scorevectheta{\B{x}} = \nabla_{\B{\theta}} \{\log \modeldist \} (\B{x})$ for OOD detection.
A naïve approach would be to measure the size of the score vector by computing the $L^2$ norm $\lVert \scorevectheta{\B{x}}\rVert_2^2$ of the gradient \cite{nguyen2019gee}. 
In the following, we analyse this idea, demonstrating its limitations: 
We empirically find that the size of the norm of the score vector is dominated by specific neural network layers which the overall gradient norm cannot capture.
In Fig. \ref{gradientHistograms}, we train deep generative models (here: Glow \cite{kingma2018glow} and diffusion models \cite{ho2020denoising}) on a training dataset (here: \texttt{CelebA}). 
We then draw a batch of items from different evaluation datasets and compute the squared \textit{layer-wise} $L^2$-norm of the gradients of the log-likelihood of a deep generative model with respect to the parameters $\B{\theta}_j$ of the corresponding layer, i.e. $f_{\B{\theta}_{j}}(\B{x}_{1:B}) = \norm{
    \nabla_{\B{\theta}_j} (\sum_{b=1}^B l(\B{x}_b))}_2^2$.
The histrogrammes in the left two columns plot $f_{\B{\theta}_{j}}(\B{x}_{1:B})$ for each layer separately, the plots in the rightmost column shows their interaction in a scatterplot.

Two points are worth noting:
We observe that for a given neural network layer (and different batches), the gradients are of a similar size, but \emph{across} layers, the scale of the layer-wise gradient norms differs by orders of magnitudes. 
In particular, taking the norm over the entire score vector would overshadow the signal of layers with a smaller norm by those on much larger magnitudes.  %
Second, note that the layer-wise gradient norms do \textit{not} strongly correlate for randomly selected layers. 
In particular, one may find two layers with corresponding features $f_{\B{\theta}_{j}}(\B{x}_{1:B})$ and $f_{\B{\theta}_{k}}(\B{x}_{1:B})$ which allow us to separate training (in-distribution) from evaluation (OOD) data points with a line in this latent space.
Considering an example, when fixing $f_{\B{\theta}_{1}}(\B{x}_{1:B}) \approx -7.5$, large negative values of $f_{\B{\theta}_{2}}(\B{x}_{1:B})$ are in-distribution, and as they become more positive, they correspond to the out-of-distribution datasets \texttt{CIFAR-10} and \texttt{ImageNet32}, respectively, with very high probability.
This renders the layer-wise information superior over the overall gradient for use as discriminative features in OOD detection.
We present our complete results in Appendix \ref{app:Additional experimental details and results} Figs. \ref{fig: gradientHistograms_app1}-\ref{fig: gradientHistograms_app3}, showing further layers (histogrammes), also for other deep generative models (VAEs \cite{kingma2014vae}) and training datasets (\texttt{SVHN}, \texttt{CelebA}, \texttt{GTSRB}, \texttt{CIFAR-10} \& \texttt{ImageNet32}).

\subsection{The Fisher Information Metric: A principled way of measuring the size of the gradient}

\label{sec: FIM definition}

Having identified the limitations of using the $L^2$ norm of the gradient, a perhaps mathematically more natural way to measure the score vector's size is to use the norm induced by the \textit{Fisher information metric} $\norm{\scorevectheta{\B{x}}}_{FIM}$ \citep{rao_1948}, defined as
\begin{align}
    \norm{\scorevectheta{\B{x}}}_{FIM}^2 = \scorevectheta{\B{x}} ^T F_{\B{\theta}}^{-1} \scorevectheta{\B{x}}, \hspace{1cm} F_{\B{\theta}} = E_{\B{y} \sim \modeldist} (\scorevectheta{\B{y}} \scorevectheta{\B{y}}^T)  \label{eq:fim}
\end{align}
where $F_{\B{\theta}}$ is called the \textit{Fisher Information Matrix (FIM)}.
Intuitively, the FIM re-scales the gradients to give more equal weighting to the parameters which typically have smaller gradients, and thus the Fisher information metric accounts for and is independent of how the parameters are scaled.  %
This in theory prevents a dependence on representation in the gradient space.

The value $\norm{\scorevectheta{\B{x}}}_{FIM}^2$ is called the score statistic, which Rao (1948) showed to follow a $\chi^2$ distribution with $\vert \B{\theta} \vert$ degrees of freedom, assuming the model parameters $\B{\theta}$ are maximum likelihood estimates. 
Thus the score statistic can be used with a statistical test known as the \emph{score test} \cite{rao_1948}.
Deep generative models with a likelihood-based objective perform a form of approximate maximum-likelihood estimation.

\begin{figure}[t]
    \centering
    \includegraphics[width=0.6\textwidth]{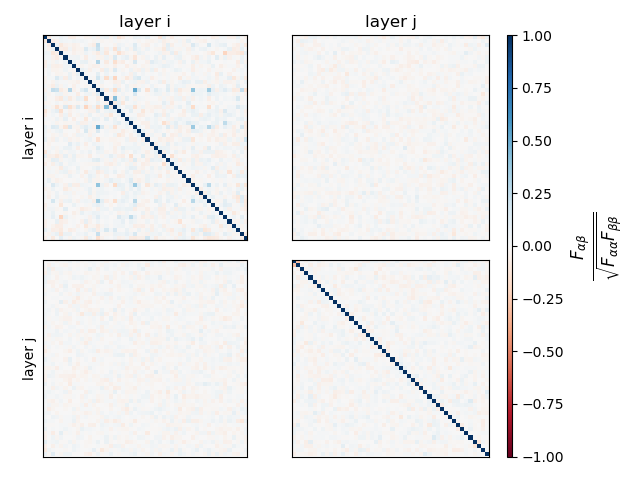}
    \caption{\textit{The layer-wise FIM has large absolute diagonal values}. We randomly select two layers $\theta_i$ and $\theta_j$ from a Glow model trained on \texttt{CelebA}, and randomly select $\max(50, \lvert \theta_j \rvert )$ weights from each layer. We then compute slices of the FIM using the method described in Equation \eqref{eq:mc_FIM_approx} and plot the results, with dark blue colours at coordinates $(\alpha, \beta)$ corresponding to larger values for the corresponding element of the FIM. 
    In order to maintain visual fidelity of the plot when weights between layers vary by orders of magnitudes, we normalise row $\alpha$ by a factor of $\sqrt{F_{\alpha \alpha}}$ where $F_{\alpha \alpha}$ indicates the element of the FIM at coordinates $(\alpha, \alpha)$, and likewise for the columns, which could be equivalently formulated as re-scaling the model parameters by this factor. 
    The same plots using diffusion models and of the raw values $F_{\alpha\beta}$ (without row and column-wise normalisation) are presented in Appendix \ref{app:FIM additional plots}, Figures \ref{fig:raw FIM glow} \& \ref{fig:raw FIM diffusion}.
    }
    \label{fig:FIM windows glow}
\end{figure}

\subsection{Approximating the Fisher Information Matrix}

\label{sec: FIM approximation}

In practise, the full FIM has $P \times P$ entries, where $P = \lvert \B{\theta} \rvert$ is the number of parameters of the deep generative model. This means that is too large to store in memory, and would furthermore be too expensive to compute and invert. For example our glow implementation has $P \approx 44 \text{ million}$ parameters, and thus the FIM would require $\approx 7,700$ terabytes to store using a \texttt{float32} representation. 
To develop a computable method, we therefore need to find a way to approximate the FIM. What would be a good choice for this approximation? --
This problem is non-trivial due to its dimensionality.  
We start answering this question by computing the FIM in Eq. \eqref{eq:fim} restricted to a subset of parameters in two layers with parameters $\B{\theta}_1, \B{\theta}_2$ of a Glow \cite{kingma2018glow} model trained on \texttt{CelebA}, using the Monte-Carlo (MC) approximation\footnote{Note that our choice to use a MC approximation is just for the sake of being able to compute $F_{\B{\theta}}$; we here do not make any further (and more principled) approximations or assumptions.}
\begin{align}
    F_{\B{\theta}_j} = E_{\B{y} \sim \modeldist} (\scorevecthetalayer{\B{y}} \scorevecthetalayer{\B{y}}^T) \approx  \frac{1}{N}\sum_{i=1}^N \scorevecthetalayer{\B{y}^{(i)}} \scorevecthetalayer{\B{y}^{(i)}}^T, 
    \qquad \B{y}^{(i)} \sim  \modeldist,  
    \label{eq:mc_FIM_approx}
\end{align}
where $\nabla_{\B{\theta}_j}$ refers to taking the gradient with respect to the parameters $\B{\theta}_j$ in layer $j$ and $\B{y}^{(i)}$ are samples drawn from the generative model $\modeldist$.
This computation is infeasible for larger layers of the network which may be highly informative for OOD detection, demonstrating the need for a more principled approximation technique.

In Fig. \ref{fig:FIM windows glow} we illustrate the resulting (restricted) FIM estimate for two layers with $N=1024$.
Further layers of this and other models, and when trained on other datasets are presented in Appendix \ref{app:FIM additional plots}, Figures \ref{fig:FIM windows glow} - \ref{fig:raw FIM diffusion}.
We observe an interesting pattern of \textit{diagonal dominance}: 
The diagonal elements are significantly larger in absolute value, on average approximately five times the size of the off-diagonal elements. 
Hence, a seemingly `crude', yet as turns out highly efficient approximation of the layer-wise FIM is to approximate as a multiple of the identity matrix, which corresponds to computing the \textit{layer-wise} $L^2$ norm of the gradient.
This reduces the cost from inverting an arbitrary $P \times P$ with potentially large P to approximating one value for each of the layers $\B{\theta}_1, \B{\theta}_2 \dots \B{\theta}_J$ of the model.

\subsection{A method for exploiting layer-wise gradients}

\label{sec: method description}

We are now interested in operationalising our observations so far into an algorithm that can be used in practice. 

In addition to the diagonal dominance phenomenon enabling a layer-wise approximation via a diagonal matrix, recall that layer-wise gradients contain more information than the overall gradient norm as the scale of the gradient norms differs by orders of magnitudes. We are therefore motivated to consider each layer $\B{\theta}_1, \B{\theta}_2 \dots \B{\theta}_J$ in our model separately and combine the results as an OOD score in the second step. 
Specifically, if we select a layer $\B{\theta}_j$ we can consider a restricted model where only the parameters in this layer are variable, and the other layers are frozen. 
We can approximate the score statistic (\ref{eq:fim}) on this restricted model, whose parameters are more homogeneous in nature. 
In practice, we take the score vector for a layer $\nabla_{\B{\theta}_j} l(\B{x})$ and attempt to approximate $\norm{\nabla_{\B{\theta}_j} l(\B{x})}_{FIM}^2$, which should follow a $\chi^2$ test with $\vert \B{\theta}_j \vert$ degrees of freedom for in-distribution data.
As discussed in \S \ref{sec: FIM approximation}, we approximate the FIM restricted to this layer as a multiple of the identity. 
For a batch of $B$ (possibly equal to $1$) of data points $\B{x}_{1:B}$ we define features $f_{\B{\theta}_j}$, which via our identity-matrix approximation should be proportional to $\norm{\nabla_{\B{\theta}_j} l(\B{x}_{1:B})}_{FIM}^2$, by
\begin{align}
    f_{\B{\theta}_j}(\B{x}_{1:B}) =
    \norm{
        \nabla_{\B{\theta}_j} \left(\sum_{b=1}^{B} l(\B{x}_b))\right)
    }_2^2. \label{eq:feature_definition}
\end{align}
Given that these layer-wise $L^2$ norms $f_{\B{\theta}_j}$ should be proportional to a $\chi^2$ distributed variable with a large degree of freedom, we expect $\log f_{\B{\theta}_j}$ to be normal-distributed \cite{bartletts1946chisquare}. 
In Fig. \ref{gradientHistograms} (further results in Appendix \ref{app:Additional experimental details and results}) we observe a good fit of $\log f_{\B{\theta}_j}$ to a Normal distribution, empirically validating this holds in spite of our approximations. %

\begin{figure}[t]
    \centering
    \includegraphics[width=0.45\textwidth]{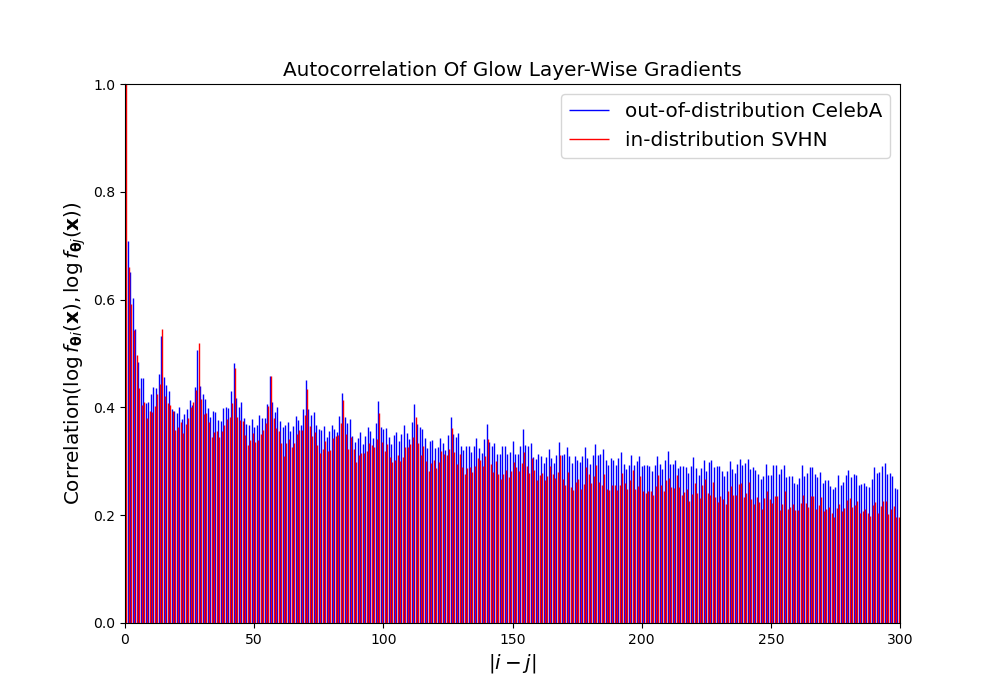}
    \includegraphics[width=0.45\textwidth]{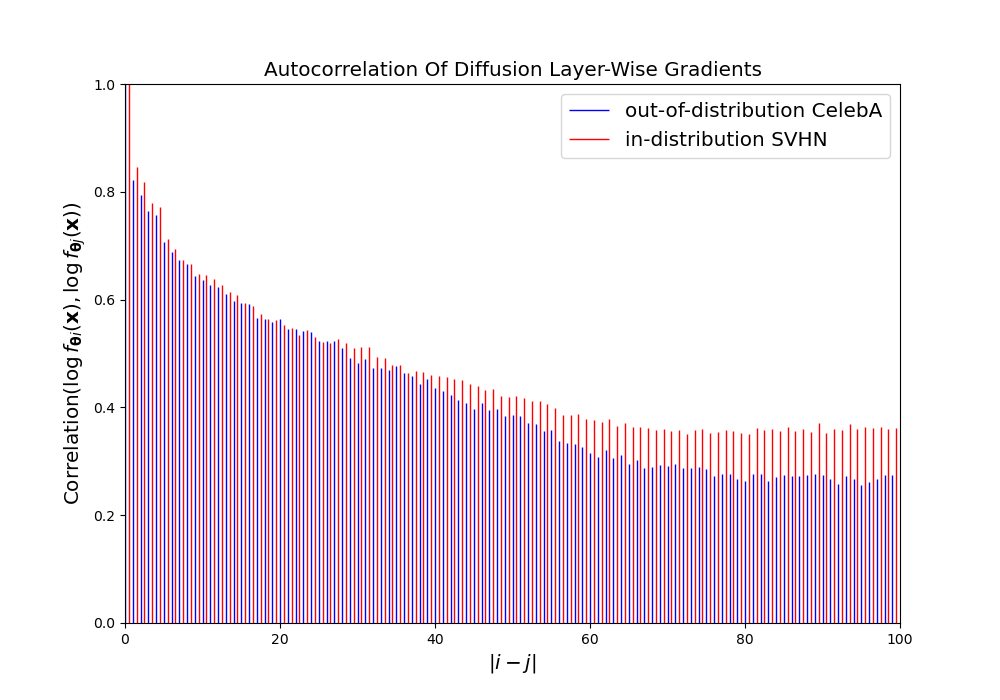}
    \caption{\textit{The $L^2$-norms of layer-wise gradients have little correlation.} We select layers with parameters $\B{\theta}_i, \B{\theta}_j$ and measure the correlation of the logarithm gradient $L^2$-norms $\log f_{\B{\theta}_{i}}(\B{x})$. 
    Binning these correlations by the distance between the layers $\vert i - j\vert$ and averaging across correlations of this distance gives the above plot. 
    We note that there is a strong correlation in $L^2$-norm between adjacent layers, but that this correlation quickly decays for both in-distribution and out-of-distribution data. 
    We hypothesise that this enables our approximation of the FIM which assumes independence across layers to provide good performance.}
    \label{fig:gradient autocorrelation}
\end{figure}

This gives rise to a natural method of combining the layer-wise $L^2$ norms: 
we simply fit Normal distributions to each log-feature $\log f_{\B{\theta}_j}$ independently, and then use the joint density as an ``in-distribution'' score. 
Algorithms \ref{alg: feature computation} to \ref{alg: anomaly deteciton} summarise our proposed method. 
As with other unsupervised OOD detection methods \citep{nalisnick2019}, we assume the existence of a small fit set of data held-out during training to accurately fit to each feature $f_j$.
Note that in practise our method is very straight-forward to implement, requiring only a few lines of PyTorch code. 
Many other methods could possibly be constructed from our theoretical and empirical insights, and we will discuss potential future work in \S \ref{sec:Conclusion}.  

In Appendix \ref{app:fisher's method comparison} we observe a mild performance improvement, uniformly across datasets, with the joint density approach in comparison to using Fisher's method \cite{Fisher_method_1938} when combining these statistics using z-scores. 
We hypothesise that this could be due to the density being more robust to correlation between adjacent layers as noted in Fig. \ref{fig:gradient autocorrelation}. 
Our presented method does not enjoy the full invariance under rescaling of the model parameters as the true score statistic (see \S \ref{sec: FIM definition}).
However, in Appendix \ref{proof:parameterisationInvar} we show that it does satisfy invariance when rescaling each layer individually, justifying our use of the density in this setting. 
Our method satisfies the desiderata of the principle of invariance (see \S \ref{sec: related work}), is hyperparameter-free, and is applicable to any data modality and model with a differentiable estimate of the log-likelihood.

\begin{algorithm}[h!]
    \caption{Algorithm for computing features}
    \label{featureFunc}
    
    \begin{algorithmic}

        \Require Deep generative model $M$, with parameters $\B{\theta}_1, \B{\theta}_2, \dots \B{\theta}_J$ in each of its $J$ layers.

        \State

        \Function {gradient features}{$\B{x}_1 \dots \B{x}_B$}

        \For{$\B{x_b}$ in batch}
            \State $l(\B{x}_b) \gets M(\B{x}_b)$
            \Comment{Compute the log-likelihood}
        \EndFor
        
        \State $v_{\B{\theta}} \gets \nabla_{\B{\theta}} (l(\B{x}_1) + \dots + l(\B{x}_B))$ 
        \Comment{Compute the gradient via backpropagation}

        \For{$j \gets 1 \dots J$ in layers}
            \State $f_j \gets 
            \norm{v_{\B{\theta}_j}}_2^2$
            \Comment{Store the layer-wise $L^2$ norms}
        \EndFor
            
        \EndFunction
    \end{algorithmic}
    \label{alg: feature computation}

\end{algorithm}

\begin{algorithm}[h!]
    \caption{Algorithm for training models}
    \label{trainingAlgo}
    \begin{algorithmic}
        \Require train dataset and held-out fit dataset 
        \State Train a deep generative model $M$, with parameters $\B{\theta}_1, \B{\theta}_2, \dots \B{\theta}_J$ in each of its $J$ layers.
        \For{Batch $\B{x}_1^n \dots\B{x}_B^n $ in fit dataset}
             \State $f_1^n \dots f_J^n \gets  \Call{gradient features}{\B{x}_1^n \dots\B{x}_B^n}$ 
        \EndFor
        \For{$j \gets 1 \dots J$ in layers}
            \State $\mu_j \gets \Call{mean}{\log f_j^1 \dots \log f_j^N}$
            \State $\sigma_j^2 \gets \Call{variance}{\log f_j^1 \dots \log f_j^N}$
            \Comment{Fit Gaussians to logarithmic features}
        \EndFor
    \end{algorithmic}

    \label{alg: model fitting}

\end{algorithm}

\begin{algorithm}[h!]
    \caption{Algorithm for detecting OOD data}
    \label{testingAlgo}
    \begin{algorithmic}
        \State Given new batch of samples $\B{y}_1 \dots \B{y}_B$
        \State $\B{f} \gets \Call{gradient features}{\B{y}_1 \dots \B{y}_B}$
        \State 
         $S(\B{y}_1 \dots \B{y}_B) = - \log \mathcal{N}(\log \B{f}; \B{\mu} ; \text{Diag}(\B{\sigma}^2))$
        \Comment{Set OOD score to be the Gaussian negative log likelihood}
    \end{algorithmic}
    \label{alg: anomaly deteciton}

\end{algorithm}

\subsection{Application to diffusion models}

A denoising diffusion model \cite{pmlr-v37-sohl-dickstein15, ho2020denoising} uses a chain of latent variables $\{\B{x}_n\}_{t = 0}^{t = T}$ achieved by gradually adding noise (represented by the conditional distributions $q(\B{x}_{t} \vert \B{x}_{t-1})$) to the initial distribution of images $\B{x} = \B{x}_0$ until the final latent variable $\B{x}_T$ is approximately Gaussian with mean zero and identity variance. The inverse process is then learned by a model approximating $p^{\B{\theta}}(\B{x}_{t-1} \vert \B{x}_{t})$. 
Diffusion models have gained in popularity due to their ability to produce samples with high visual fidelity and semantic coherence, making them a natural candidate for OOD detection \cite{Graham_2023_CVPR}. 
Nonetheless, they do not allow for exact inference of the log-likelihood, only a variational lower bound thereof, and this variational lower bound is very expensive to compute as it requires running $\vert T \vert$ inference steps on the network modelling $p^{\B{\theta}}(\B{x}_{t-1} \vert \B{x}_{t})$. For our setup, $\vert T \vert = 1000$ and running a full-forward/backward pass requires roughly $1$ minute of compute per sample. 
Thus, we choose one component of the variational lower bound, namely the one-step log-likelihood,

\[
    l(\B{x}) = l(\B{x}_0) = \mathbb{E}_{\B{x}_1 \sim q(\B{x}_1 \vert \B{x}_0)}\log p^{\B{\theta}}(\B{x}_0 \vert \B{x}_1)
\]

We refer to Appendix \ref{app:diffusion ablation tables} for computational details, noting that our implementation using one sample from $q(\B{x}_1 \vert \B{x}_0)$ only requires $1$ pass on of inference on $p^{\B{\theta}}(\B{x}_{0} \vert \B{x}_{1})$. 
Despite this value being very different to the intractable full log-likelihood $p^{\B{\theta}}(\B{x}_{0})$, in Figure \ref{likelihoodHistogram} we observe the same open problem and phenomenon as \cite{nalisnick2018deep} reported for the full likelihood estimates of other deep generative models. 
In Appendix \ref{app:diffusion ablation tables} we perform an ablation study on this one-step component of the variational lower bound used, finding the result that for both our method and the typicality method \cite{nalisnick2019}, the components which are computed with less noise added to the image $\B{x}_t$ used as input to the model $p^{\B{\theta}}(\B{x}_{t-1} \vert \B{x}_{t})$ are more informative for OOD detection, which could intuitively be understood as the noised image $\B{x}_t$ itself being more informative in this regard.

\section{Related work}

\label{sec: related work}

In this section we review related work which uses gradients for unsupervised OOD detection.

In concurrent work to ours, \cite{choi2021robust, bergamin2022model} each present an approximation to Rao's score test \citep{rao_1948}. 
They independently approached the problem from the directions of training on the given sample of OOD data \cite{zhisheng2020regret} and application of tests from classical statistics, respectively. 
These methods use approximations of the FIM from the field of optimization \cite{amari1998naturalgrad, tieleman2012rmsprop, kingma2015adam}, whereas we use a simpler approximation tailored to the task of unsupervised OOD detection, and complement this with our empirical observations of the FIM in \S \ref{sec: FIM approximation}. 
\cite{bergamin2022model} compute a score test across the whole model by approximating the FIM as a diagonal, with elements $F_{\alpha \alpha} = (\partial_{\alpha} \log p^{\B{\theta}} (\B{x}))^2 + \epsilon$ for a small hyperparameter $\epsilon = 10^{-8}$, which which is used in optimization for its damping effect \cite{martens2020naturalgrad} and helps to mitigate numerical instabilities when dividing by $F_{\alpha \alpha}$. 
Our method differs in that it explicitly encodes the layer-homogeneity of the model (whereby parameters in the same layer have similar gradient sizes and perform similar functions in the network), and the predicted chi-square distribution of the score.
We also note that layers whose gradient values are $<< 10^{-8}$ (see Appendix \ref{app:FIM additional plots} Figures \ref{fig:raw FIM glow} and \ref{fig:raw FIM diffusion}) would have their information nullified without careful tuning of $\epsilon$. 
\cite{choi2021robust} split the problem of OOD detection layer-wise, but use the more complex EKFAC \cite{thomas2018ekfac} algorithm to account for dependencies between adjacent parameters. 
After some normalisation and additional processing steps, the authors compute the ROSE metric by taking the maximum feature over some pre-selected subset of layers. 
Our method differs as it uses a holistic score influenced by all the model layers.

\cite{nguyen2019gee} are interested in using VAEs to detect anomalous web traffic in a semi-supervised setting, measuring the difference between a test gradient and labelled anomalous gradients.
Our method differs in that does not require anomalous examples.
\cite{kwon2020backpropagated} computes a cosine similarity between the gradients in the decoder of a VAE and the average gradients observed during training as their OOD metric of choice. 
In this work, we advocate for using the \textit{size} of the gradient vector rather than its angle as done in \cite{kwon2020backpropagated}: 
our intuition is that, for a well-trained model evaluated on in-distribution data, we are close to a local minimum where the gradient is flat and the variance of the angle of the gradient vector is high. 
In particular, when averaging over samples from the model, we have that $\mathbb{E}_{\B{x} \sim \modeldist }\nabla_{\B{\theta}}  (\log \modeldist ) (\B{x}) = \B{0}$ as a distribution minimises its own cross-entropy. 
In their supplementary, \cite{nalisnick2019} note that the Maximum Mean and Kernelized Stein Discrepancy tests they use to benchmark their typicality test are only effective when using the Fisher kernel \cite{jaakkola1998exploiting, jacot2018neural} $k(x_i, x_j) = \nabla_{\theta} l(x_i) \nabla_{\theta} l(x_j)^T$. 
Our method differs from using Fisher kernel in that it allows for information to be used from all the layers, rather than a few dominant layers, as we discuss in \S \ref{sec:Layer-wise gradients are highly informative and differ in size by orders of magnitudes}. 

To the best of our knowledge, no previous work has connected these works bar \cite{bergamin2022model}'s citation of \cite{choi2021robust}. The theoretical grounding which we provide in Proposition \ref{prop:gradient_invariance} \S \ref{sec: proof_gradient_invariance} may explain why multiple other authors have independently found efficacy in unsupervised OOD detection with gradient information.

For completeness, in Appendix \ref{sec: classifier gradient methodology} we review previous work on the use of gradients of classifiers for the task of supervised OOD detection.

\section{Experimental benchmark}
\label{sec: Experimental Results}

In this section, we benchmark our OOD detection method against the typicality test. 
We postpone detailed description of our datasets and models to Appendix \ref{app:Code, computational requirements, existing assets.}.

We follow the consensus of previous literature \cite{nalisnick2018deep} to evaluate our method on distribution pairs: training a generative image models on one image distribution and testing against a surrogate out-distribution. 
We choose five natural image datasets \texttt{SVHN}, \texttt{CelebA}, \texttt{GTSRB}, \texttt{CIFAR-10} and \texttt{ImageNet32} used in previous literature \cite{serra2019input} and evaluate on all dataset pairings. 
To the best of our knowledge this makes our evaluation more extensive than any previously published work in unsupervised OOD detection, a field where rigorous evaluation is especially important as erroneously high performance can be achieved by selective reporting of or fine-tuning hyperparameters to certain out-distributions.

\begin{table}[H]
\caption{Comparison of the AUROC values (higher is better) of our method to the typicality test \cite{nalisnick2019} for batch sizes $B = 1, 5$. 
We train Glow \cite{kingma2018glow} models on five natural image datasets (columns) and evaluate the ability of the model-method combination to reject the other datasets (rows).
Bold indicates the element-wise higher value comparing both methods. 
}
\label{tab:GlowResults}
\begin{tabular}{l | l | r r r r r}
\toprule
\multicolumn{2}{l |}{\textit{test} $\downarrow$  \textit{train} $\rightarrow$}
 & \texttt{SVHN} & \texttt{CelebA} & \texttt{GTSRB} & \texttt{CIFAR-10} & \texttt{ImageNet32} \\
\midrule
\multirow[c]{5}{*}{\rotatebox{90}{\parbox{1.5cm}{typicality \\ $(B = 1)$ }}}
 & \texttt{SVHN} & - & 0.8735 & 0.3469 & 0.8599 & \textbf{0.8915} \\
 & \texttt{CelebA} & \textbf{0.9989} & - & 0.6506 & 0.3680 & 0.2857 \\
 & \texttt{GTSRB} & 0.9261 & 0.8201 & - & 0.6708 & 0.5548 \\
 & \texttt{CIFAR-10} & \textbf{0.9829} & 0.7733 & 0.6423 & - & 0.4147 \\
 & \texttt{ImageNet32} & 0.9952 & 0.9251 & 0.8057 & 0.7249 & - \\
\cline{1-7}
\multirow[c]{5}{*}{\rotatebox{90}{\parbox{1.5cm}{ours \\ $(B = 1)$ }}} 
 & \texttt{SVHN} & - & \textbf{0.9880} & \textbf{0.9858} & \textbf{0.8747} & 0.8010 \\
 & \texttt{CelebA} & 0.9823 & - & \textbf{0.9262} & \textbf{0.5155} & \textbf{0.2997} \\
 & \texttt{GTSRB} & \textbf{0.9537} & \textbf{1.0000} & - & \textbf{0.7546} & \textbf{0.9967} \\
 & \texttt{CIFAR-10} & 0.9658 & \textbf{0.9462} & \textbf{0.9126} & - & \textbf{0.4377} \\
 & \texttt{ImageNet32} & \textbf{0.9976} & \textbf{0.9876} & \textbf{0.9683} & \textbf{0.7375} & - \\
\midrule
\midrule
\multirow[c]{5}{*}{\rotatebox{90}{\parbox{1.5cm}{typicality \\ $(B = 5)$ }}}
 & \texttt{SVHN} & - & 0.9899 & 0.6119 & 0.9961 & \textbf{0.9983} \\
 & \texttt{CelebA} & \textbf{1.0000} & - & 0.9786 & 0.4737 & 0.4293 \\
 & \texttt{GTSRB} & \textbf{0.9997} & 0.8987 & - & 0.6639 & 0.6138 \\
 & \texttt{CIFAR-10} & \textbf{1.0000} & 0.9082 & 0.9613 & - & 0.4894 \\
 & \texttt{ImageNet32} & \textbf{1.0000} & 0.9974 & 0.9954 & 0.9013 & - \\
\cline{1-7}
\multirow[c]{5}{*}{\rotatebox{90}{\parbox{1.5cm}{ours \\ $(B = 5)$ }}}
 & \texttt{SVHN} & - & \textbf{0.9997} & \textbf{1.0000} & \textbf{0.9989} & 0.9976 \\
 & \texttt{CelebA} & 0.9997 & - & \textbf{1.0000} & \textbf{0.9525} & \textbf{0.8514} \\
 & \texttt{GTSRB} & 0.9996 & \textbf{0.9999} & - & \textbf{0.9596} & \textbf{0.9999} \\
 & \texttt{CIFAR-10} & 0.9992 & \textbf{0.9970} & \textbf{1.0000} & - & \textbf{0.6712} \\
 & \texttt{ImageNet32} & \textbf{1.0000} & \textbf{0.9995} & \textbf{1.0000} & \textbf{0.9480} & - \\
\bottomrule
\end{tabular}
\end{table}

In Tables \ref{tab:GlowResults} \& \ref{tab:DiffusionResults} we compare our method against the typicality test \cite{nalisnick2019} using the Area Under Receiver Operating Curve (AUROC) statistic on both single sample $(B=1)$ batch size $(B=5)$ OOD detection. 
We choose typicality as it is, to the best of our knowledge, the most performant method which is both model-agnostic and hyper parameter free. 
Performance of unsupervised OOD detection can vary greatly depending on the model and even the image resizing algorithm applied make the inputs of uniform size \cite{bergamin2022model}. 
To mitigate this problem we directly compare to our implmentation of \cite{nalisnick2019} using the same models and dataset implementations where they exist.

For Glow models (Table \ref{tab:GlowResults}) our method outperforms typicality on most dataset pairings, whereas for diffusion models (Table  \ref{tab:DiffusionResults}) neither method dominates, although our method achieves higher average AUROC. 
We hypothesise that the comparative advantage our method enjoys for Glow models could be related to the model having more layers (1353 vs. 276) leading to more gradient features being available. 
In Appendix \ref{sec:VAEresults} table \ref{tab:VAEresults} we note poor performance for both methods when applied to a VAE model with poor sample quality, indicating that how well the model captures the dataset is the main factor driving performance of the downstream OOD detection method. 
Single sample $(B=1)$ performance for both methods was lower for models trained on the semantically diverse datasets \texttt{CIFAR-10} and \texttt{ImageNet32}. 
We would like to question the implicit assumption made by prior works that an unsupervised method trained on these datasets \emph{should} consistently reject images from other natural image datasets.
There is no meaningful semantic boundary to distinguish an image from \texttt{CIFAR-10} and \texttt{ImageNet32}.

As noted in \S \ref{sec: related work} our method is similar to those presented in concurrent works \cite{choi2021robust, bergamin2022model}, we choose to use our method as a representative of this class of methods so that we may use our compute resources to robustly investigate their performance over such a wide range of distribution pairings and models. 
We make no claim of superior performance over these methods.

\begin{table}[H]
\caption{
\textit{Diffusion models} Comparison of the AUROC values (larger values are better) of our method to the typicality test \cite{nalisnick2019} for batch sizes $B = 1, 5$. We train Diffusion \cite{ho2020denoising} models on five natural image datasets (as columns) and evaluate the ability of the model-method combination to reject the other datasets (as rows).
Bold indicates the element-wise higher value comparing both methods. 
}
\label{tab:DiffusionResults}
\begin{tabular}{l | l | r r r r r}
\toprule
\multicolumn{2}{l |}{\textit{test} $\downarrow$  \textit{train} $\rightarrow$}
 & \texttt{SVHN} & \texttt{CelebA} & \texttt{GTSRB} & \texttt{CIFAR-10} & \texttt{ImageNet32} \\
\midrule
\multirow[c]{5}{*}{\rotatebox{90}{\parbox{1.5cm}{typicality \\ $(B = 1)$ }}}
 & \texttt{SVHN} & - & 0.9357 & 0.4661 & \textbf{0.9007} & \textbf{0.8777} \\
 & \texttt{CelebA} & \textbf{0.9990} & - & 0.3860 & 0.3409 & 0.2837 \\
 & \texttt{GTSRB} & \textbf{0.9335} & 0.8197 & - & \textbf{0.6981} & \textbf{0.5624} \\
 & \texttt{CIFAR-10} & \textbf{0.9920} & 0.6968 & 0.4855 & - & 0.4142 \\
 & \texttt{ImageNet32} & \textbf{0.9986} & 0.8759 & \textbf{0.6759} & \textbf{0.7443} & - \\
\cline{1-7}
\multirow[c]{5}{*}{\rotatebox{90}{\parbox{1.5cm}{ours \\ $(B = 1)$ }}} 
 & \texttt{SVHN} & - & \textbf{0.9903} & \textbf{0.8526} & 0.5574 & 0.6214 \\
 & \texttt{CelebA} & 0.9551 & - & \textbf{0.5466} & \textbf{0.5655} & \textbf{0.3571} \\
 & \texttt{GTSRB} & 0.8691 & \textbf{0.9684} & - & 0.5622 & 0.5530 \\
 & \texttt{CIFAR-10} & 0.9535 & \textbf{0.9639} & \textbf{0.5786} & - & \textbf{0.4710} \\
 & \texttt{ImageNet32} & 0.9363 & \textbf{0.9818} & 0.6651 & 0.5763 & - \\
\midrule
\midrule
\multirow[c]{5}{*}{\rotatebox{90}{\parbox{1.5cm}{typicality \\ $(B = 5)$ }}}
 & \texttt{SVHN} & - & 0.9978 & 0.7943 & \textbf{0.9975} & \textbf{0.9961} \\
 & \texttt{CelebA} & \textbf{1.0000} & - & 0.7642 & 0.3156 & 0.3621 \\
 & \texttt{GTSRB} & \textbf{0.9998} & 0.8336 & - & 0.6809 & 0.5765 \\
 & \texttt{CIFAR-10} & \textbf{1.0000} & 0.7808 & \textbf{0.8332} & - & 0.4488 \\
 & \texttt{ImageNet32} & \textbf{1.0000} & 0.9866 & \textbf{0.9675} & \textbf{0.9266} & - \\
\cline{1-7}
\multirow[c]{5}{*}{\rotatebox{90}{\parbox{1.5cm}{ours \\ $(B = 5)$ }}}
 & \texttt{SVHN} & - & \textbf{1.0000} & \textbf{0.9970} & 0.8457 & 0.9561 \\
 & \texttt{CelebA} & 0.9908 & - & \textbf{0.8552} & \textbf{0.7734} & \textbf{0.4202} \\
 & \texttt{GTSRB} & 0.9716 & \textbf{0.9997} & - & \textbf{0.7325} & \textbf{0.9007} \\
 & \texttt{CIFAR-10} & 0.9895 & \textbf{0.9992} & 0.8104 & - & \textbf{0.5733} \\
 & \texttt{ImageNet32} & 0.9862 & \textbf{1.0000} & 0.9309 & 0.8532 & - \\
\bottomrule
\end{tabular}
\end{table}

\section{Conclusion} 
\label{sec:Conclusion}

We analysed an approximation to the Fisher information metric for OOD detection.
Our work has two key limitations: 
First, while we have provided the most extensive empirical benchmark of deep generative models, OOD and in-distribution datasets, datasets beyond images and for instance large language models should be tested. 
Second, while we focused on comparing it to the best performing, model-agnostic, hyperparameter-free OOD method, further empirical benchmarking against other methods should be conducted.
Future work should investigate other, potentially more computationally expensive methods for approximating the Fisher information metric and its use in OOD detection.

\bibliography{bib.bib}
\bibliographystyle{tmlr}

\newpage

\appendix

{\Large \textbf{Appendix for \papertitle}}

\input{A1_theory_proofs}

\newpage

\input{A2_AddExperimentalDetailsResults}

\newpage
\input{A3_background}

\newpage

\input{A4_datasets}
\newpage

\input{A5_negSoc}
\newpage
\input{A6_code}

\newpage
\input{A11_VAE_results}
\newpage

\end{document}

%% file: math_commands.tex
\usepackage{amsmath,amsfonts,bm}

\def\eqref#1{equation~\ref{#1}}

\def\1{\bm{1}}

\DeclareMathAlphabet{\mathsfit}{\encodingdefault}{\sfdefault}{m}{sl}
\SetMathAlphabet{\mathsfit}{bold}{\encodingdefault}{\sfdefault}{bx}{n}

\newcommand{\E}{\mathbb{E}}

\newcommand{\R}{\mathbb{R}}

%% file: 2_draftSwitch.tex
\newif\ifdraft
\drafttrue   %

%% file: _utility.tex
\usepackage[utf8]{inputenc} %
\usepackage[T1]{fontenc}    %
\usepackage{url}            %
\usepackage{booktabs}       %
\usepackage{amsfonts}       %
\usepackage{nicefrac}       %
\usepackage{microtype}      %
\usepackage{multicol}
\usepackage{subfiles}
\usepackage[hyperfootnotes=true, hidelinks]{hyperref}
\hypersetup{ %
    pdftitle={},
    pdfauthor={},
    colorlinks=true,
    linkcolor=MidnightBlue,
    citecolor=MidnightBlue,
    filecolor=MidnightBlue,
    urlcolor=MidnightBlue,
    }
\usepackage{soul} %
\usepackage{bbding}
\usepackage{pifont}
\usepackage{caption}

\makeatletter
\newif\ifnobrackets
\renewcommand\@cite[2]{\ifnobrackets\else[\fi{#1\if@tempswa , #2\fi}\ifnobrackets\else]\fi\nobracketsfalse}

\makeatother

\usepackage{url}            %
\PassOptionsToPackage{hyphens}{url}
\hypersetup{
    allcolors = MidnightBlue,
}

\usepackage{booktabs}       %
\usepackage{amsfonts}       %
\usepackage{amsmath}
\usepackage{amssymb}
\usepackage{nicefrac}       %
\usepackage{microtype}      %
\usepackage{wrapfig}
\usepackage{sidecap}

\usepackage{bm}
\usepackage{amsthm}

\usepackage{tikz}
\usetikzlibrary{bayesnet}
\usepackage{graphicx}
\usepackage{caption}
\usepackage{subcaption}
\usepackage{paralist}

\usepackage{algorithmicx}
\usepackage{algorithm}
\usepackage{algpseudocode}

\usepackage{array,multirow,graphicx}
\usepackage{float}

\usepackage{wrapfig}

\usepackage{todonotes}
\usepackage{tikz-cd}
\usepackage{bbm}
\usepackage[shortlabels,inline]{enumitem}

\usepackage{multicol}

\usepackage{wrapfig}

\usepackage{lscape}

\newcommand{\norm}[1]{ \left\| #1 \right\| }

\theoremstyle{definition}

\newtheorem{remark}{Remark}
\newtheorem{prop}{Proposition}
\newtheorem*{theorem*}{Theorem}

\newcommand*\colourcross[1]{%
  \expandafter\newcommand\csname #1cross\endcsname{\textcolor{#1}{\ding{55}}}%
}
\colourcross{black}

\newif\ifdraft
\draftfalse %

\usepackage{todonotes}
\usepackage{tikz-cd}
\usepackage{bbm}

\usepackage{booktabs}

\definecolor{MidnightBlue}{rgb}{0.1, 0.1, 0.44}

\newcommand{\B}[1]{\boldsymbol{#1}}

\usepackage{letltxmacro}
\renewcommand{\eqref}[1]{(\ref{#1})}

%% file: A1_theory_proofs.tex
\section{Proofs and additional theoretical results}
\label{app:Proofs and additional theoretical results}

\subsection{Proof of proposition \ref{prop:gradient_invariance}}

\label{sec: proof_gradient_invariance}

\paragraph{Proposition 1.} Let $\modeldist_{\mathcal{X}}(\B{x})$ and $\modeldist_{\mathcal{T}}(\B{t})$ be two probability density functions corresponding to the same model distribution $\modeldist$ being represented on two different measure spaces $\mathcal{X}$ and $\mathcal{T}$. Suppose these representations encode the same information, i.e. there exists a smooth, invertible reparameterization $T: \mathcal{X} \rightarrow \mathcal{T}$ such that for $\B{x} \in \mathcal{X}$ and $\B{t} \in \mathcal{T}$ representing the same point we have $T(\B{x}) = \B{t}$.
Then, the gradient vector $\nabla_{\B{\theta}} ( \log \modeldist)$ is invariant to the choice of representation, and in particular, 
$\nabla_{\B{\theta}} ( \log \modeldist_{\mathcal{T}} )( \B{t}) 
=
\nabla_{\B{\theta}}  (\log \modeldist_{\mathcal{X}} ) (\B{x})$.

\paragraph{Proof.} Via the change-of-variables formula, we obtain

\begin{align*}
    \modeldist_{\mathcal{T}}(\B{t}) = \modeldist_{\mathcal{X}}(\B{x}) 
    \; \left\lvert 
    \frac{\partial T^{-1} }{\partial \B{x}} 
    \right\rvert.
\end{align*}

Applying the logarithm on both sides provides
\begin{align*}
    \log \modeldist_{\mathcal{T}}(\B{t}) = \log \modeldist_{\mathcal{X}}(\B{x}) +
    \log
    \; \left\lvert 
    \frac{\partial T^{-1} }{\partial \B{x}} 
    \right\rvert,
\end{align*}

and hence $\nabla_{\B{\theta}} ( \log \modeldist_{\mathcal{T}} )( \B{t}) = \nabla_{\B{\theta}}  (\log \modeldist_{\mathcal{X}} ) (\B{x})$ as required.

The smoothness assumption could be relaxed by considering the pull-back measure $\mathbb{P}^{\B{\theta}}_{\mathcal{X}} \circ T^{-1} = \mathbb{P}^{\B{\theta}}_{\mathcal{T}}$ and the corresponding change-of-variables formula for Radon-Nikodym derivatives, however we omit this for brevity and relevance. This result also trivially extends to the likelihood proxy we use for diffusion models $\log \modeldist (\B{x}_0 \vert \B{x}_1)$.

\subsection{Representation-Invariance of variatonal lower-bounds}

\label{elboInvar}

Assume the same setup as in \ref{sec: proof_gradient_invariance}, but this time with a variational Bayesian method such as a Variational AutoEncoder \cite{kingma2014vae} with latent variable given by $\B{z}$, decoder probability density $p^{\B{\theta}}_{\mathcal{X}}( \B{x} \lvert \B{z})$ and encoder probability density $q^{\B{\phi}}(\B{z} \lvert \B{x})$, noting that the decoder probability density is that which depends on $\mathcal{X}$. The Evidence Lower Bound on the log-likelihood $p^{\B{\theta}}_{\mathcal{X}}( \B{x} )$ is given by
\begin{align*}
    ELBO^{\B{\theta}, \B{\phi}}_{\mathcal{X}}(\B{x}) = 
    \mathbb{E}_{\B{z} \sim q^{\B{\phi}}(\B{z} \lvert \B{x})} 
    \left(
        \log
        \frac{p^{\B{\theta}}_{\mathcal{X}}(\B{x}, \B{z})}
             {q^{\B{\phi}}(\B{z} \lvert \B{x})}
    \right).
\end{align*}
We can then state a similar representation invariance for the ELBO.

\begin{prop}
    Let $ELBO^{\B{\theta}, \B{\phi}}_{\mathcal{X}}$ be the ELBO of a VAE, and let $ELBO^{\B{\theta}, \B{\phi}}_{\mathcal{T}}(\B{t})$ be the ELBO under a change of variables with invertible mapping $T: \mathcal{X} \rightarrow \mathcal{T}$, corresponding to two sets $\mathcal{X}$ and $\mathcal{T}$.
    Then, $ELBO^{\B{\theta}, \B{\phi}}_{\mathcal{T}}(\B{t})$ is invariant to $T$.
\end{prop}

\textbf{Proof. }
Noting that $\quad p^{\B{\theta}}_{\mathcal{T}}(\B{t}, \B{z}) = p^{\B{\theta}}_{\mathcal{X}}(\B{x}, \B{z}) 
\left\lvert 
    \frac{\partial T^{-1} }{\partial \B{x}} 
\right\rvert \quad$ 
while 
$\quad q^{\B{\phi}}(\B{z} \lvert \B{t}) = q^{\B{\phi}}(\B{z} \lvert \B{x}) \quad $ 
\footnote{To those familiar with the Borel–Kolmogorov paradox this condition may seem non-obvious, but we can derive it from the fact that $T$ does not require input from $\B{z}$, and thus 
$q^{\B{\phi}}(\B{z} \lvert \B{t}) = 
\frac{q^{\B{\phi}}(\B{z}, \B{t}) }{q^{\B{\phi}}(\B{t})} = 
\frac{q^{\B{\phi}}(\B{z}, \B{x}) \left\lvert \frac{\partial T^{-1} }{\partial \B{x}} \right\rvert}
{q^{\B{\phi}}(\B{x}) \left\lvert \frac{\partial T^{-1} }{\partial \B{x}} \right\rvert} = 
q^{\B{\phi}}(\B{z} \lvert \B{x})$}
gives that:
\begin{align*}
     ELBO^{\B{\theta}, \B{\phi}}_{\mathcal{T}}(\B{t}) = ELBO^{\B{\theta}, \B{\phi}}_{\mathcal{X}}(\B{x}) + \log \left\lvert 
        \frac{\partial T^{-1} }{\partial \B{x}}
    \right\rvert
\end{align*}
and taking the gradient wrt $\B{\theta}$ and $\B{\phi}$ gives the result that the gradient of the ELBO with respect to the VAE's parameters is representation-invariant.

\subsection{Lebesgue measure of a set of bounded total variation}

\label{proof:volumeTotalVariation}

\begin{prop}

For $\B{x} \in \R^d$, define the total variation to be $TV(\B{x}) = \lvert x_1 \rvert + \sum_{i=2}^{d} \lvert x_{i} - x_{i-1} \rvert$. Let $E(\alpha)$ be the set of $d$-length arrays whose total variation is bounded by $\alpha$:

\[
    E(\alpha) = \{\B{x} \in \R^d: TV(\B{x})\}.
\]

The Lebesgue measure of this set is given by $E(\alpha) = \frac{(2\alpha)^d}{\Gamma(d+1)}$.

\end{prop}

\textbf{Proof. }

Consider the volume-preserving transformation $(x_1, x_2 \dots x_d) \mapsto (x_1, t_2 \dots t_d)$, where $t_i = x_i - x_{i-1}$. We thus see that the volume of $E(\alpha)$ is equivalent to the volume of the $d$-ball in the $\ell^1$-metric, with a standard result:

\[
    \mu(E(\alpha)) = \mu( \{(x_1, t_2 \dots t_d): 
    \lvert x_1 \rvert +  \lvert t_2 \rvert + \dots +  \lvert t_d \rvert\})
    =
    \frac{(2\alpha)^d}{\Gamma(d+1)}.
\]

\textbf{Application to MNIST} We can na\"ively apply this result to MNIST images $\B{y} \in [0, 1]^{28 \times 28}$ by setting $d = 28^2$ and drawing a snake pattern through our images, setting $y_{ij} = x_{28(j - 1) + (-1)^{j + 1}(i - 14) + 14}$. Computing this numerically for the whole MNIST dataset, we see that $\alpha = 102.9$ is sufficiently large such that the whole MNIST dataset is contained in $E(\alpha)$, which we can compute has an approximate measure of $E(\alpha) \approx 10^{-116.76} \leq 10^{-116}$. Note that this is not the tightest bound one could give; for example vertical variations are neglected and membership in $E(\alpha)$ does not restrict $x_{i}$ from drifting outside the set $[0, 1]$. %

\subsection{(Weak) parameterisation in-variance of our method}

\label{proof:parameterisationInvar}

Let $\Theta, \Phi$ be two parameter spaces of the same model $p$, linked by the smooth invertible reparameterisation $P: \Theta \to \Phi$, such that for $\B{\phi} = P(\B{\theta})$ we have $p^{\B{\theta}} = p^{\B{\phi}}$. In this setting, one can derive that the Fisher Information Metric \cite{rao_1948} is invariant under $P$, ie that for all $\B{x}_1, \B{x}_2 \in \mathcal{X}$ we have $\nabla_{\B{\theta}} l (\B{x}_1) F_{\B{\theta}}^{-1} \nabla_{\B{\theta}} l (\B{x}_2)^T = \nabla_{\B{\phi}} l (\B{x}_1) F_{\B{\phi}}^{-1} \nabla_{\B{\phi}} l (\B{x}_2)^T$ (see (\ref{eq:fim}) for our notation). As we merely approximate the FIM in our method we cannot make the same guarantee for all $P$, we can however prove a similar result if $P$ linearly rescales the layers:

\begin{prop}
    As in \S \ref{sec: method description} Let $\B{\theta}_1, \B{\theta}_2 \dots \B{\theta_J}$ be the layers of our model. $P: \Theta \to \Phi$ be a smooth invertible reparameterisation of our model which linearly rescales the layers, ie $P(\B{\theta}_1, \B{\theta}_2 \dots \B{\theta}_J) = d_1 \B{\theta}_1, d_2 \B{\theta}_2 \dots d_J \B{\theta}_J$ for some constants $d_1, d_2 \dots d_Jv\in \R$. Then, the resulting anomaly score of our method is invariant under $P$.
\end{prop}

\textbf{Proof. }

Using the same notation as in \S \ref{sec: method description}, let $f_1^{\Theta} \dots f_J^{\Theta}$ and  $f_1^{\Phi} \dots f_J^{\Phi}$  be our layer-wise gradient $L^2$ norm features under $\Theta$ and $\Phi$ respectively (see equation \ref{eq:feature_definition}). Then, for all datapoints $\B{x}$ and layers $j$ we have:

\begin{align}
    f_j^{\Theta} (\B{x})
    = 
    \norm{\nabla_{\B{\theta}_j} l(\B{x})}^2
    =
    \norm{d_j \nabla_{\B{\phi}_j} l(\B{x})}^2
    =
    d_j^2
    f_j^{\Phi} (\B{x})
    \label{eq: f_j reparam}.
\end{align}

Taking the logarithm and writing in vectorized form gives that:

\begin{align}
    \log \B{f}^{\Theta} (\B{x}) = 2 \log \B{d} + \log \B{f}^{\Phi}  (\B{x})
\end{align}

In particular, if we let $\B{\mu}^{\Theta}, \B{\mu}^{\Phi}$ and $\B{\sigma}^{2\Theta}, \B{\sigma}^{2 \Phi}$ be the corresponding sample mean and variances for $\Theta$ and $\Phi$ in algorithm \ref{alg: anomaly deteciton}, we see that $\B{\mu}^{\Theta} = 2 \log \B{d} +  \B{\mu}^{\Phi}$  and $\B{\sigma}^{2\Theta} = \B{\sigma}^{2\Phi} = \B{\sigma}^2$. Hence via translation invariance of the normal distribution, our metric will be invariant under $P$.

%% file: A2_AddExperimentalDetailsResults.tex
\section{Additional experimental details and results}
\label{app:Additional experimental details and results}

\subsection{RGB-HSV representation dependence}
\label{app:Additional RGB-HSV}

\begin{figure}[H]
    \centering
    \includegraphics[width=0.7\textwidth]{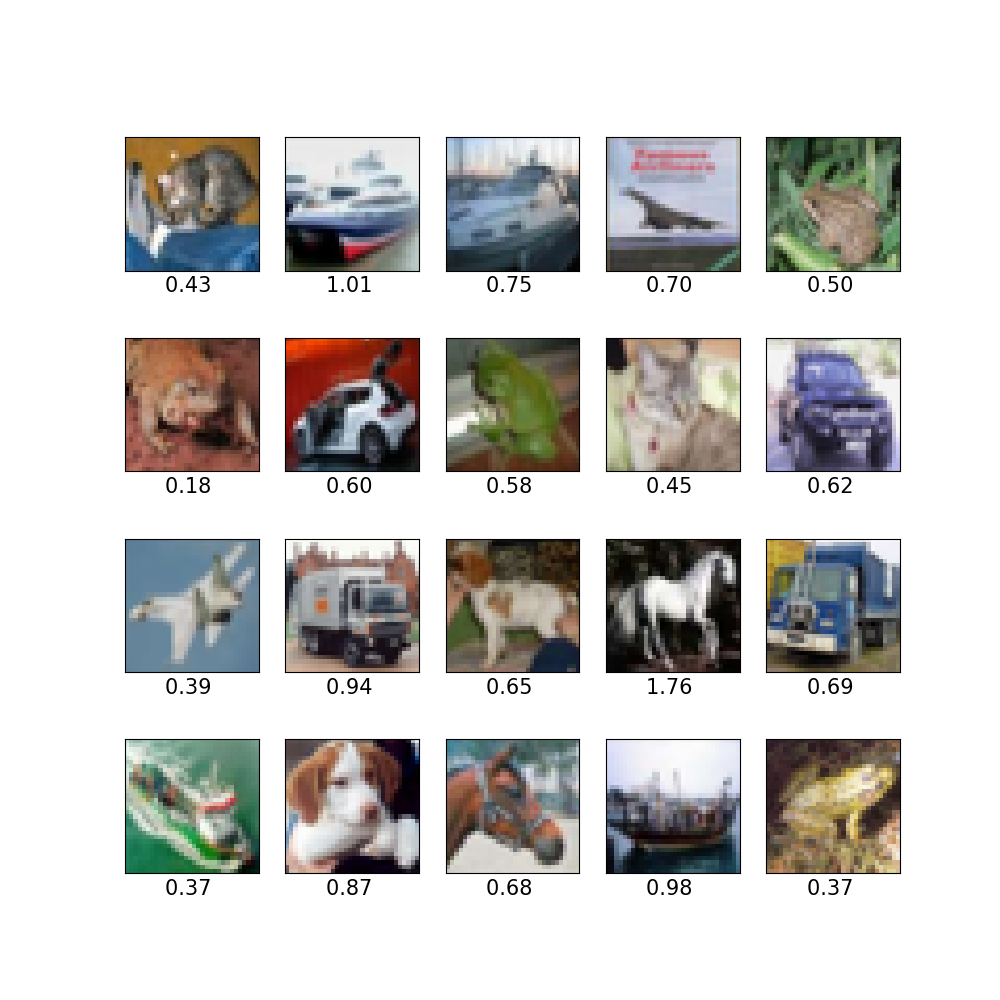}
    \caption{
    \textit{The log-likelihood heavily depends on data representation} We extend Figure \ref{fig:RGB-HSV} to
    the first $20$ examples of the \texttt{CIFAR10} dataset and their values of $\Delta^{RGB \to HSV}_{BPD}$ as defined in Eq. \ref{eq:BPD gain definition}. We note values of $\Delta^{RGB \to HSV}_{BPD}$ between $0.18$ to $1.76$, indicating a large difference of the induced change in likelihoods.}
    \label{fig:RGB-HSV appendix}
\end{figure}

Here we compute the change in Bits Per Dimension (BPD) $\Delta^{RGB \to HSV}_{BPD}$ for the first $20$ samples of the \texttt{CIFAR10} test dataset, defined as:

\begin{equation}
    \Delta^{RGB \to HSV}_{BPD} 
    = 
    \frac{1}{3 \times 32 \times 32} \log_2 \frac{d \mu_{HSV}}{d \mu_{RGB}}
    =
    \frac{\log_2 p_{RGB}(\mathbf{x}) - \log_2 p_{HSV}(\mathbf{x})} {3 \times 32 \times 32},
    \label{eq:BPD gain definition}
\end{equation}
where $\mu_{HSV}$ is the Lebesgue measure in HSV-space, $\mu_{RGB}$ is the Lebesgue measure in $RGB$-space, and $p_{HSV}$ and $p_{RGB}$ are corresponding probability density functions for any distribution defined over the set of images. We compute the Radon-Nikodym derivative $\frac{d \mu_{HSV}}{d \mu_{RGB}}$ in \ref{eq:BPD gain definition} by computing the pixel-wise Jacobian determinants of the RGB-HSV transformation $T^{RGB \to HSV}: \R^3 \to \R^3$. In order to make the comparison fair, we dequantaize pixel $x_{ij} \in \R^3$ add a small amount of normally distributed noise $\epsilon_{ij} \sim \mathcal{N}(\B{0}, \B{I}_{3 \times 3})$, ie we set $\tilde{x}_{ij} = x_{ij} + \frac{\epsilon_{ij}}{255}$. We then note that the full RGB-HSV transformation factors as a RGB-HSV transformation across the pixels, and thus its Jacobian determinant factors as:

\[
    \log \frac{d \mu_{HSV}}{d \mu_{RGB}}
    =
    \sum_{1 \leq i, j \leq 32} \log \left\vert \frac{\partial T}{\partial x_{ij}} \right\vert_{\tilde{x}_{ij}}.
\]

\newpage

\subsection{Replications of Fig. \ref{gradientHistograms}}

\begin{figure}[H]
    \centering
    \includegraphics[width=\textwidth]{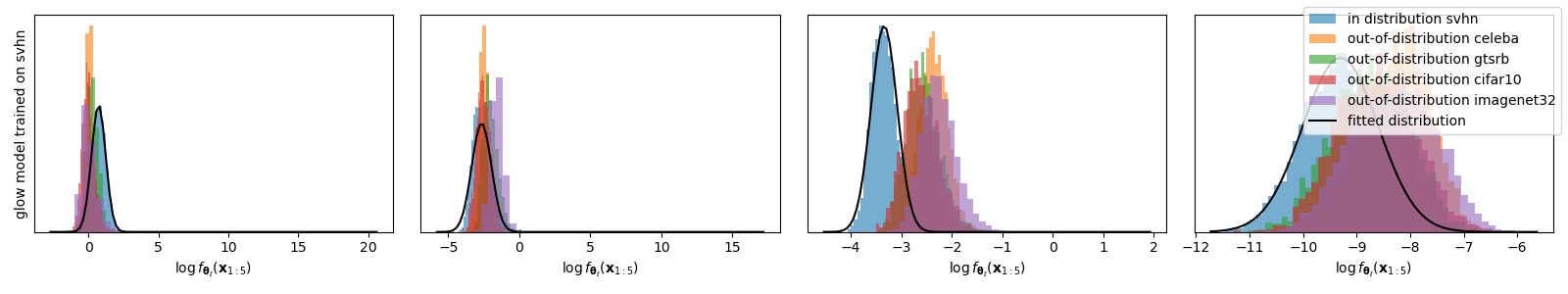}
    \includegraphics[width=\textwidth]{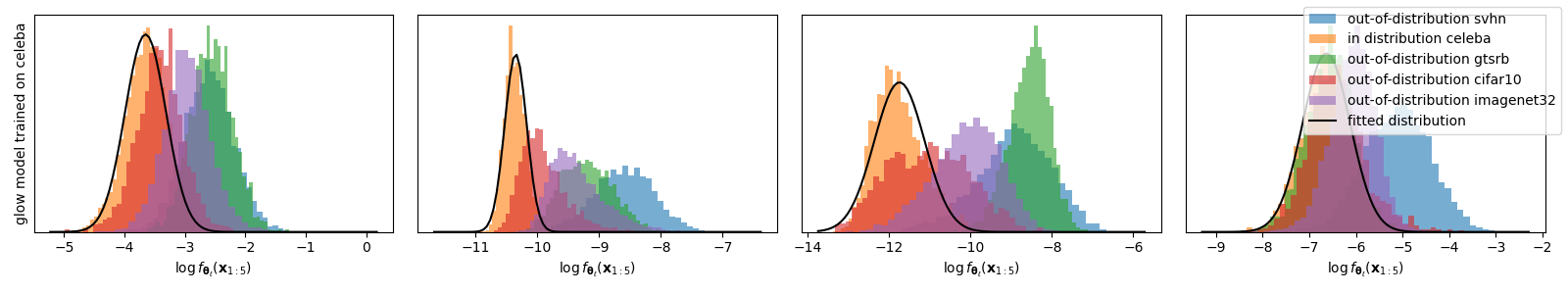}
    \includegraphics[width=\textwidth]{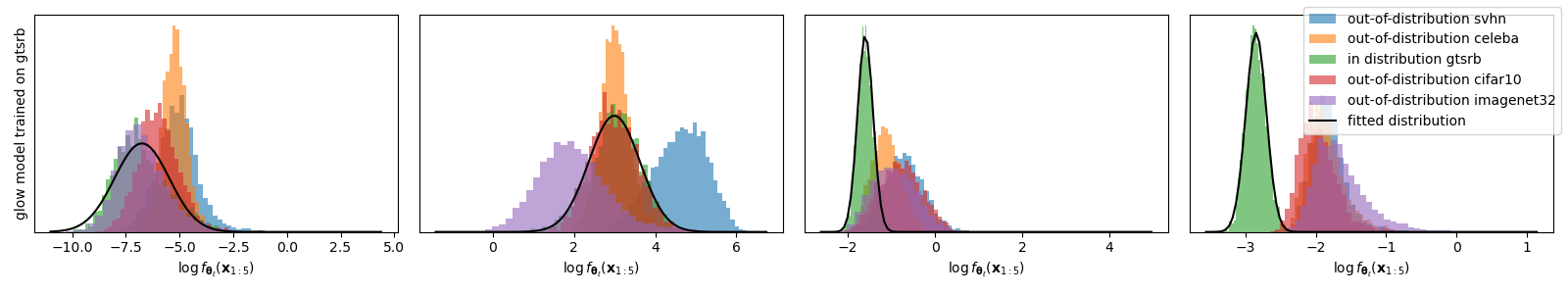}
    \includegraphics[width=\textwidth]{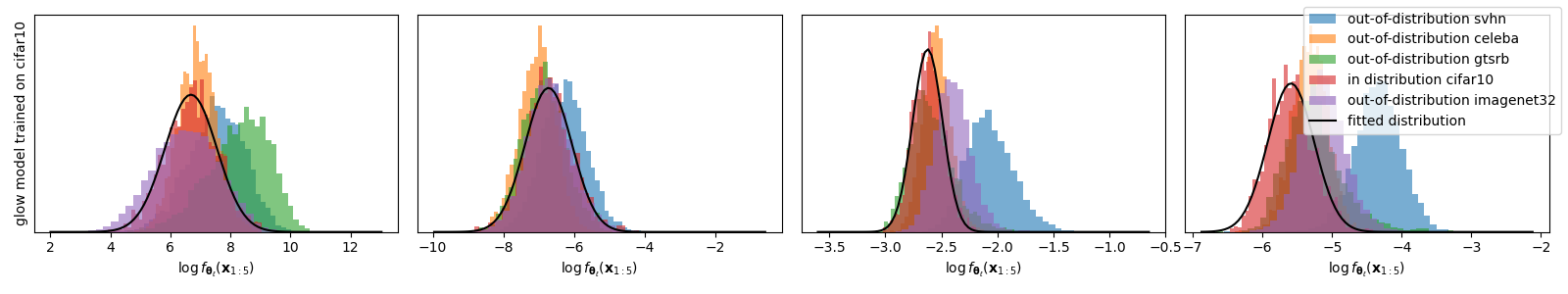}
    \includegraphics[width=\textwidth]{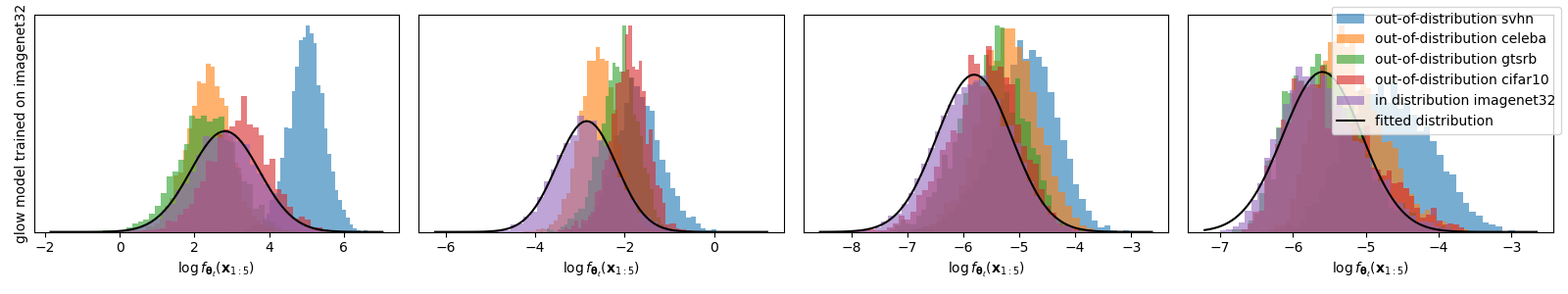}
    \caption{Replication of Fig. \ref{gradientHistograms} for $4$ randomly selected layers out of $1353$ from Glow models trained on \texttt{SVHN}, \texttt{CelebA}, \texttt{GTSRB}, \texttt{CIFAR-10} and \texttt{ImageNet32} respectively}
    \label{fig: gradientHistograms_app1}
\end{figure}

\begin{figure}[H]
    \centering
    \includegraphics[width=\textwidth]{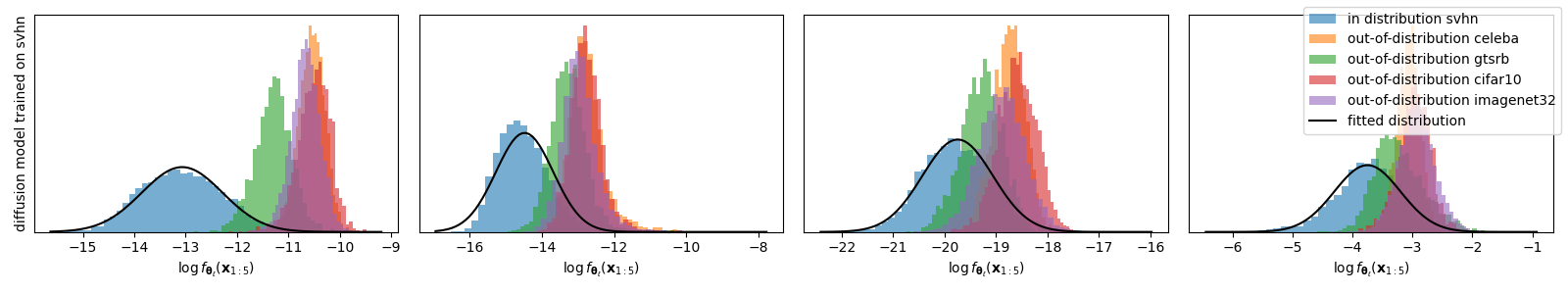}
    \includegraphics[width=\textwidth]{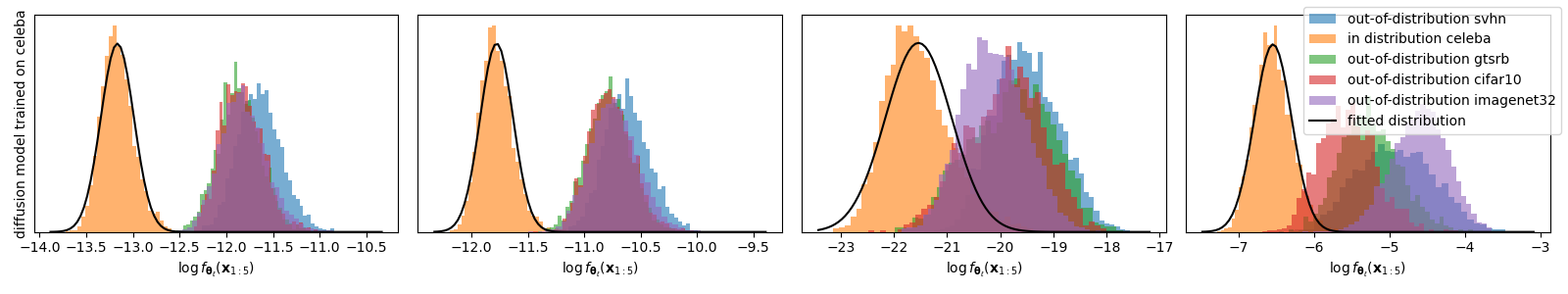}
    \includegraphics[width=\textwidth]{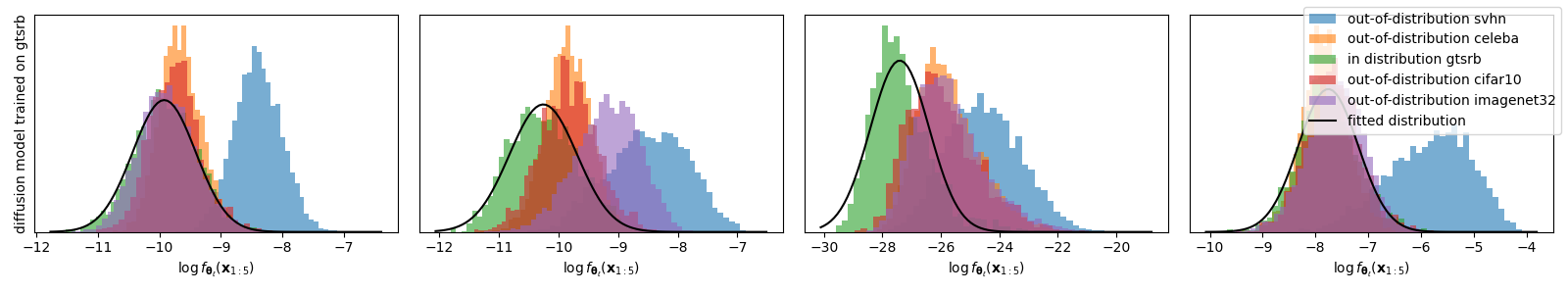}
    \includegraphics[width=\textwidth]{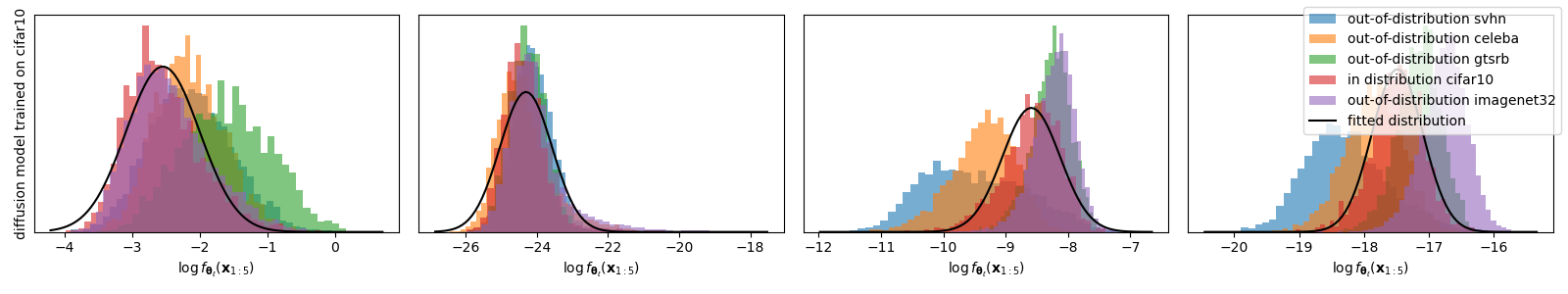}
    \includegraphics[width=\textwidth]{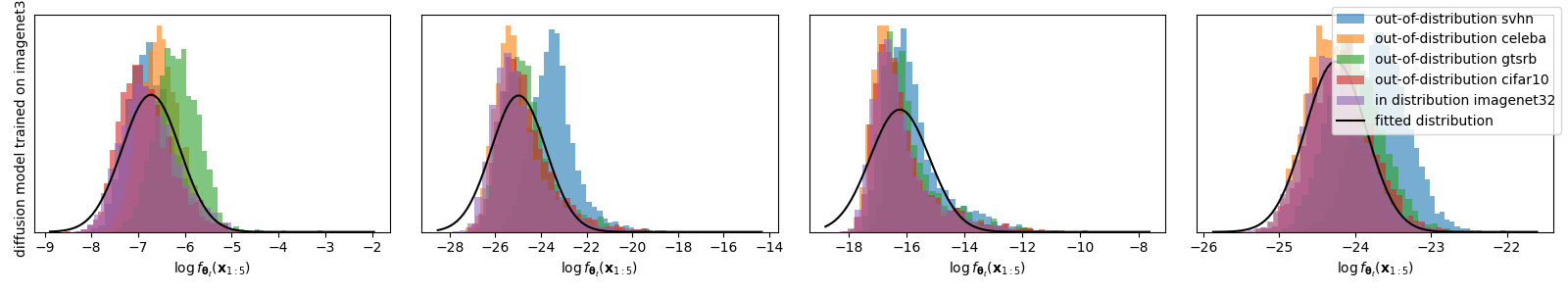}
    \caption{Replication of Fig. \ref{gradientHistograms} for $4$ randomly selected layers out of $276$ from Diffusion models trained on \texttt{SVHN}, \texttt{CelebA}, \texttt{GTSRB}, \texttt{CIFAR-10} and \texttt{ImageNet32} respectively}
    \label{fig: gradientHistograms_app2}
\end{figure}

\begin{figure}[H]
    \centering
    \includegraphics[width=\textwidth]{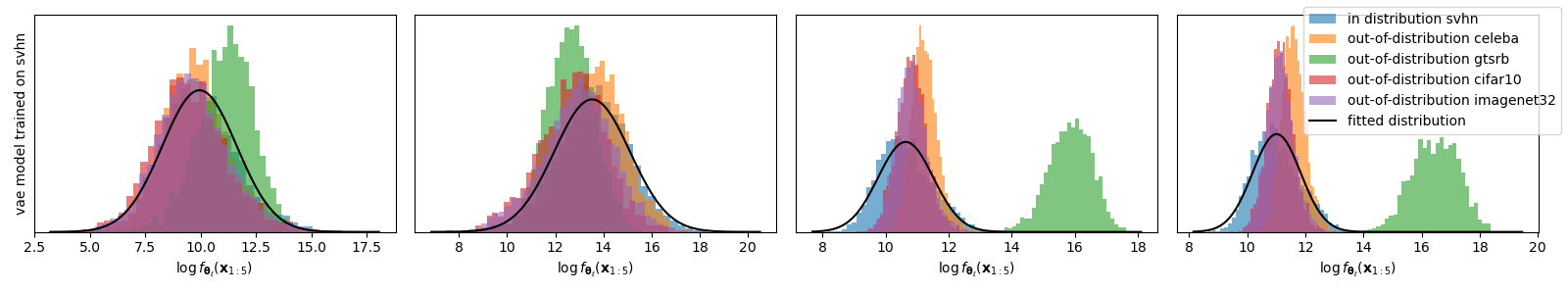}
    \includegraphics[width=\textwidth]{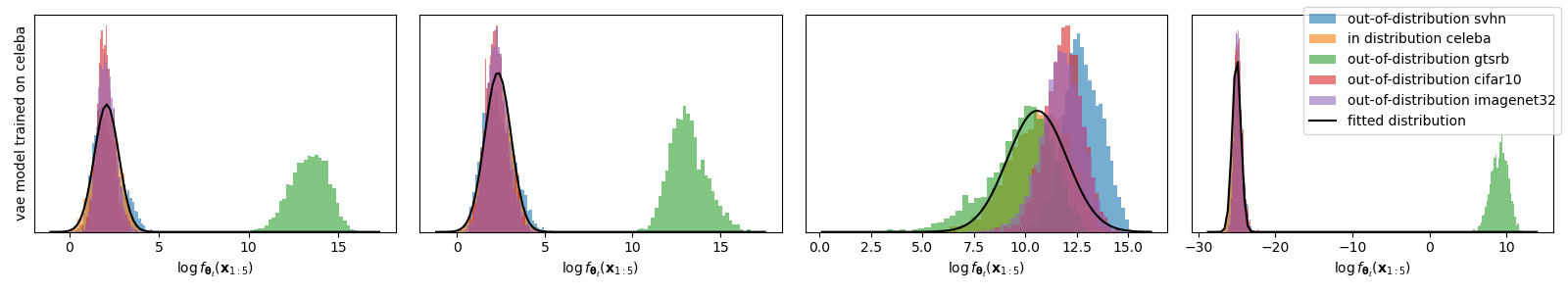}
    \includegraphics[width=\textwidth]{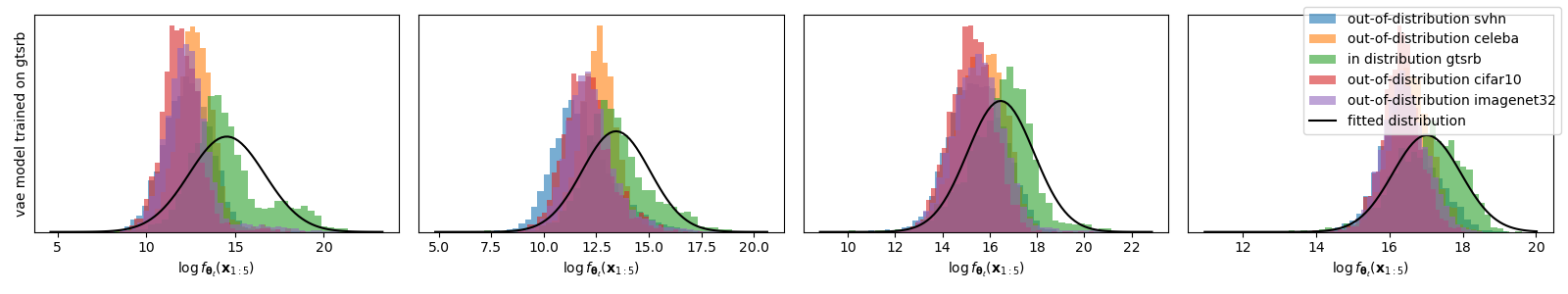}
    \includegraphics[width=\textwidth]{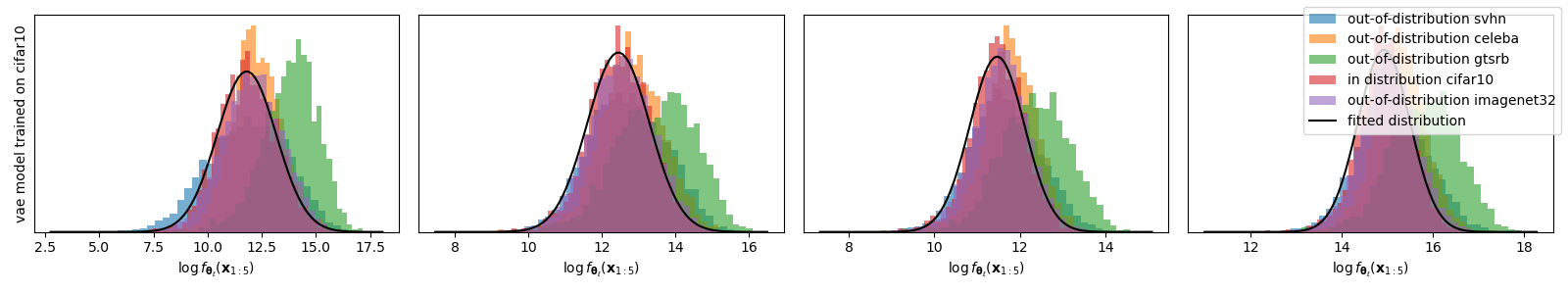}
    \includegraphics[width=\textwidth]{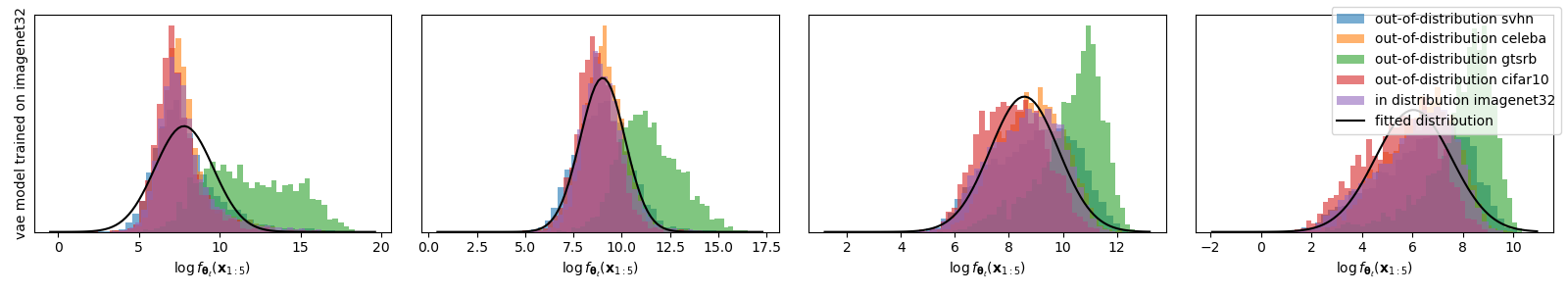}
    \caption{Replication of Fig. \ref{gradientHistograms} for $4$ randomly selected layers out of $48$ from vae models trained on \texttt{SVHN}, \texttt{CelebA}, \texttt{GTSRB}, \texttt{CIFAR-10} and \texttt{ImageNet32} respectively}
    \label{fig: gradientHistograms_app3}
\end{figure}

In figures \ref{fig: gradientHistograms_app1}- \ref{fig: gradientHistograms_app3} we provide robust replications of Figure \ref{gradientHistograms} using randomly chosen layers. The layers are sorted with the right-hand layer being the ``deepest" (ie. the closest to the latent variables). We observe that the gradients are more separated for models trained on the semantically distinct datasets \texttt{SVHN}, \texttt{CelebA} \& \texttt{GTSRB}, mirroring the superior performance our method achieves in these cases.

Please note that large parts of the gradient distributions from the OOD datasets have been cropped out to keep the plots legible.

\subsubsection{Glow models}

\newpage

\subsubsection{Diffusion models}

\newpage

\subsubsection{VAEs}

We note generally less separation with our VAE models, mirroring the poorer performance we attain with them in appendix \ref{sec:VAEresults}

\input{A8_Additional_FIM_plots}

\input{A7_diffusion_ablation_tables}

\subsubsection{Ablation study on multiple q-samples for anomaly detection with diffusion models}
\label{app: multi q sample ablation}

In this section, we investigate if any performance improvement can be achieved by using multiple samples from $q$ to estimate $L_0$, ie $l(\B{x}) = l(\B{x}_0) = \mathbb{E}_{\B{x}_1 \sim q(\B{x}_1 \vert \B{x}_0)}\log p^{\B{\theta}}(\B{x}_0 \vert \B{x}_1)$
$\E L_0 \propto \mathbf{E}_{\B{x}_1 \sim q(\B{x}_0)} \norm{\B{\epsilon} - \B{\epsilon}^{\B{\theta}}(\B{x_{1}}, 1)}$. Specifically, we take $n = 5$ q-samples $\B{\epsilon}^{(1)} \dots \B{\epsilon}^{(5)}, \B{x_{1}^{(1)}} \dots \B{x_{1}^{(5)}}$  to define our likelihood proxy as:

\[
    l(\B{x}) = \frac{1}{5}\sum_{i = 1}^5 \norm{\B{\epsilon}^{(i)} - \B{\epsilon}^{\B{\theta}}(\B{x_{1}^{(i)}}, 1)}.
\]

The AUROC values using batch size $B = 5$ for our method and typicality \cite{nalisnick2019} are in table \ref{tab: multi q sample ablation}. We note little to no performance gain for our method or typicality, motivating our use of $n=1$ q-sample in our implementation for efficiency.

\begin{table}[H]
\caption{auc values for typicality [top] and ours [bottom], batch size 5 applied to a diffusion model using $5$ q-samples \newline average performance for typicality: 0.7617, \quad 25/50/75 quantiles: 0.6062 / 0.7532 / 0.9720\newline average performance for ours: 0.8987, \quad 25/50/75 quantiles: 0.8601 / 0.9525 / 0.9794}
\begin{tabular}{l | l | r r r r r}
\toprule
 &  & \texttt{SVHN} & \texttt{CelebA} & \texttt{GTSRB} & \texttt{CIFAR-10} & \texttt{ImageNet32} \\
\midrule
\multirow[c]{5}{*}{\rotatebox[origin=c]{90}{typicality}} & \texttt{SVHN} & - & 0.9251 & 0.9668 & 0.9824 & \textbf{0.9926} \\
 & \texttt{CelebA} & \textbf{0.9981} & - & 0.6328 & 0.5312 & 0.5500 \\
 & \texttt{GTSRB} & \textbf{0.9961} & 0.6768 & - & 0.6291 & 0.4716 \\
 & \texttt{CIFAR-10} & \textbf{0.9685} & 0.7890 & 0.4130 & - & \textbf{0.7174} \\
 & \texttt{ImageNet32} & \textbf{0.9959} & 0.5954 & 0.6098 & \textbf{0.7923} & - \\
\cline{1-7}
\multirow[c]{5}{*}{\rotatebox[origin=c]{90}{ours}} & \texttt{SVHN} & - & \textbf{0.9984} & \textbf{0.9930} & \textbf{0.9873} & 0.9756 \\
 & \texttt{CelebA} & 0.8938 & - & \textbf{0.9746} & \textbf{0.8140} & \textbf{0.6952} \\
 & \texttt{GTSRB} & 0.8222 & \textbf{0.9823} & - & \textbf{0.9367} & \textbf{0.8728} \\
 & \texttt{CIFAR-10} & 0.9683 & \textbf{0.9744} & \textbf{0.8922} & - & 0.5666 \\
 & \texttt{ImageNet32} & 0.9797 & \textbf{0.9793} & \textbf{0.9188} & 0.7485 & - \\
\bottomrule
\end{tabular}
\label{tab: multi q sample ablation}
\end{table}

\subsection{Using Fisher's method in the place of density estimation}

\label{app:fisher's method comparison}

In this section, we briefly investigate the use of Fisher's method \cite{Fisher_method_1938} to compute the final anomaly score when using the gradient $L^2$-norm statistics $f^\ell$ which we define in \S \ref{sec: method description}. Specifically, if we modify our method by defining $q^\ell(\B{x}) = \min(\Phi(f^{\ell}(\B{x})), 1 - \Phi(f^{\ell}(\B{x}))$ to be the $\ell$-th p-value from a two-tailed z-test and our final anomaly score as:

\[
    S = - \sum_{\ell=1}^L \log(q^\ell(\B{x})).
\]

In table \ref{tab:fisher's method comparison} for brevity we only report our results for a Glow model with $B = 1$, but note the same pattern across all models. We note a small performance detriment across all dataset pairings from using Fisher's method, motivating our use of density estimation.

\begin{table}[H]
\caption{auc values for ours (Fisher's method) [top] and ours (density estimation) [bottom], batch size 1 applied to glow \newline average performance for Fisher's method: 0.7898, \quad 25/50/75 quantiles: 0.7004 / 0.8761 / 0.9529\newline average performance for density estimation: 0.8516, \quad 25/50/75 quantiles: 0.7894 / 0.9500 / 0.9863}
\begin{tabular}{l | l | r r r r r}
\toprule
 &  & \texttt{SVHN} & \texttt{CelebA} & \texttt{GTSRB} & \texttt{CIFAR-10} & \texttt{ImageNet32} \\
\midrule
\multirow[c]{5}{*}{\rotatebox{90}{\parbox{1.5cm}{ours \\ (Fisher)}}}  & \texttt{SVHN} & - & 0.9808 & 0.9494 & 0.8358 & 0.7514 \\
 & \texttt{CelebA} & 0.9633 & - & 0.8125 & 0.5022 & 0.2686 \\
 & \texttt{GTSRB} & 0.9321 & 0.9772 & - & 0.7016 & 0.4482 \\
 & \texttt{CIFAR-10} & 0.9398 & 0.9250 & 0.8119 & - & 0.4203 \\
 & \texttt{ImageNet32} & 0.9899 & 0.9734 & 0.9165 & 0.6969 & - \\
\cline{1-7}
\multirow[c]{5}{*}{\rotatebox{90}{\parbox{1.5cm}{ours \\ (density) }}}  & \texttt{SVHN} & - & \textbf{0.9880} & \textbf{0.9858} & \textbf{0.8747} & \textbf{0.8010} \\
 & \texttt{CelebA} & \textbf{0.9823} & - & \textbf{0.9262} & \textbf{0.5155} & \textbf{0.2997} \\
 & \texttt{GTSRB} & \textbf{0.9537} & \textbf{1.0000} & - & \textbf{0.7546} & \textbf{0.9967} \\
 & \texttt{CIFAR-10} & \textbf{0.9658} & \textbf{0.9462} & \textbf{0.9126} & - & \textbf{0.4377} \\
 & \texttt{ImageNet32} & \textbf{0.9976} & \textbf{0.9876} & \textbf{0.9683} & \textbf{0.7375} & - \\
\bottomrule
\end{tabular}
\label{tab:fisher's method comparison}
\end{table}

\begin{table}[h!]
\caption{Table of results comparing the performance of our method on for a model trained on FashionMNIST at detecting OOD grayscale images to the performances of the S-score reported in \cite{serra2019input} and Watanabe-Akaike
Information Criterion reported in \cite{choi2018waic}}
\centering
\begin{tabular}{ccccc}
    \toprule
   Method & MNIST & Omniglot\\
   \midrule
    WAIC & 0.766 &  0.796 \\
       \midrule
    S using PixelCNN++ and FLIF & 0.967 & 1.000\\
    PixelCNN Gradient norms (OneClassSVM) (ours) & 0.979 & 
    1.000 \\
    \midrule
    S using Glow and FLIF & 0.998 & 1.000 \\
    Glow Gradient norms (OneClassSVM) (ours) & 0.819 & 1.000 \\
   \bottomrule
\end{tabular}
\label{PixelCNNFahionMNISTtab}
\end{table}

%% file: A8_Additional_FIM_plots.tex
\subsection{Additional plots of the FIM}

\label{app:FIM additional plots}

\subsubsection{Windows into the FIM of a Glow model}

\begin{figure}[H]
    \centering
    \includegraphics[width=0.65\textwidth]{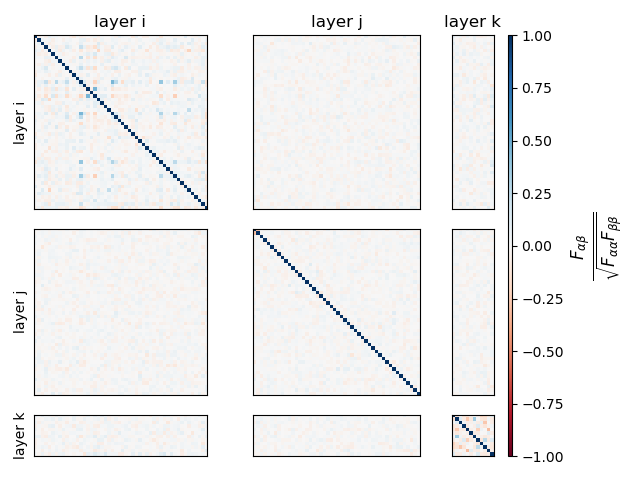}
    \caption{\textit{The strong diagonal of the FIM for glow models} we replicate Fig. \ref{fig:FIM windows glow} using three more randomly selected layers from our Glow model trained on \texttt{CelebA}.}
    \label{fig:enter-label}
\end{figure}

\subsubsection{Windows into the FIM of a diffusion model}

\begin{figure}[H]
    \centering
    \includegraphics[width=0.65\textwidth]{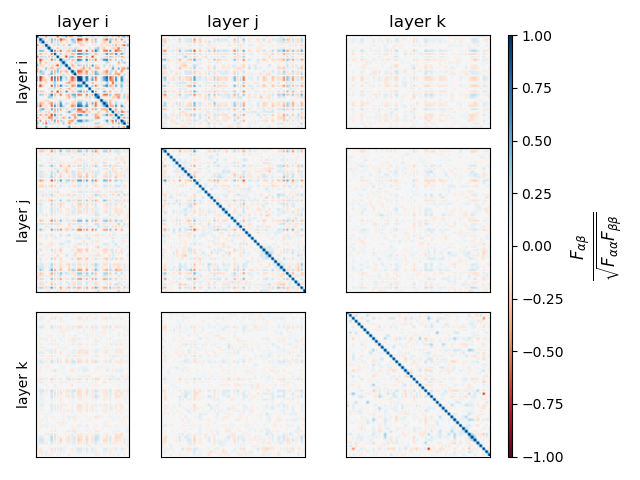}
    \caption{\textit{The strong diagonal of the FIM for diffusion models}.
    We replicate Fig. \ref{fig:FIM windows glow} with a diffusion model. As before, we randomly select two layers $\theta_\ell$ from a diffusion model trained on \texttt{CelebA} and plot the an approximation of the FIM, this time using the gradients of the one-step log-likelihood. Note that the first layer selected has fewer than $50$ weights, so we plot its entire layer-wise FIM. Again, we normalise the rows and columns by the diagonal values $\sqrt{F_{\alpha \alpha}}$ to enable cross-layer comparison.}
    \label{fig:FIM windows diffusion}
\end{figure}

\subsubsection{Raw, single-layer FIMs of Glow models}

\begin{figure}[H]
    \centering
    \includegraphics[width=0.9\textwidth]{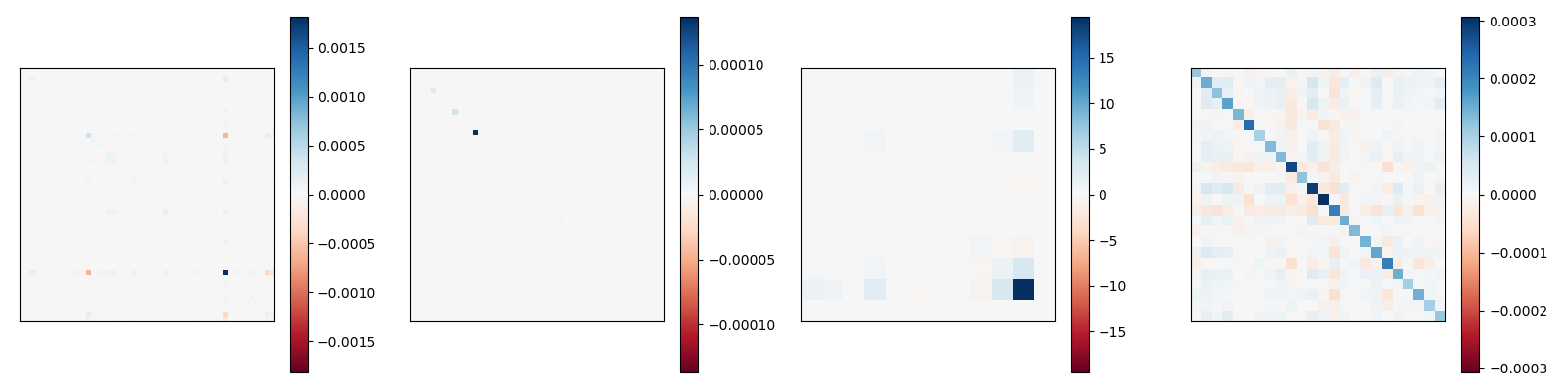}
    \includegraphics[width=0.9\textwidth]{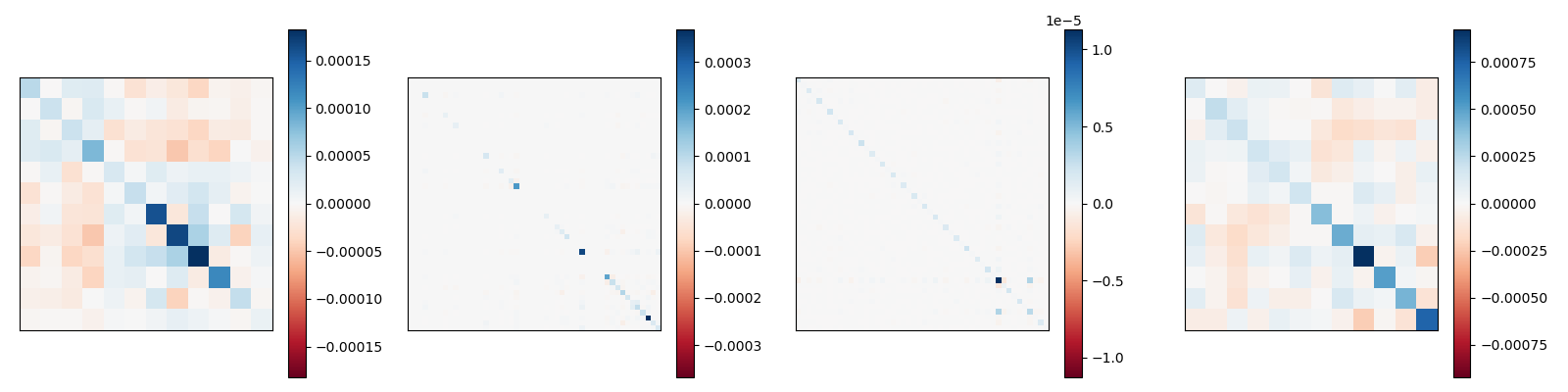}    \includegraphics[width=0.9\textwidth]{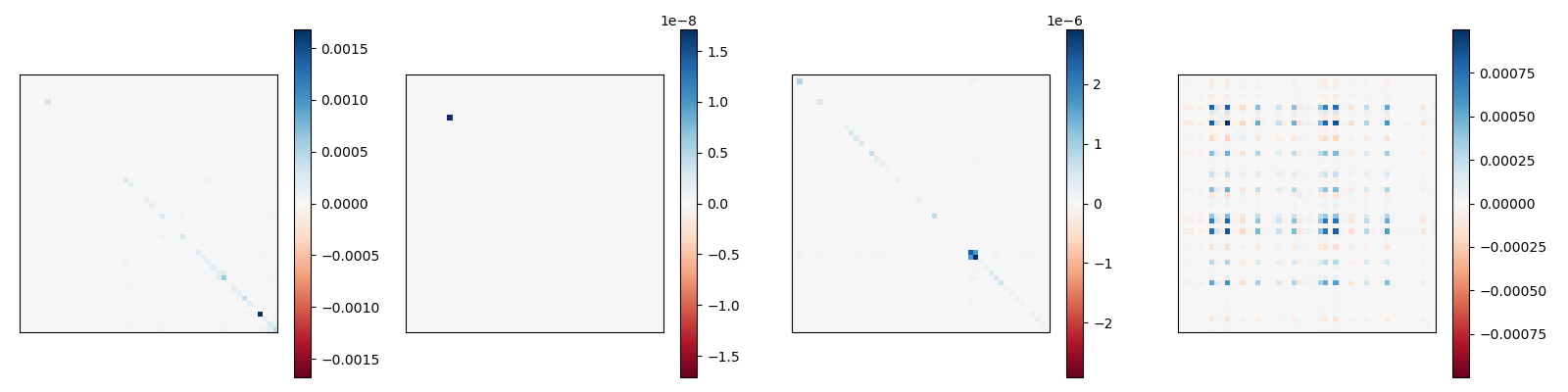}    \includegraphics[width=0.9\textwidth]{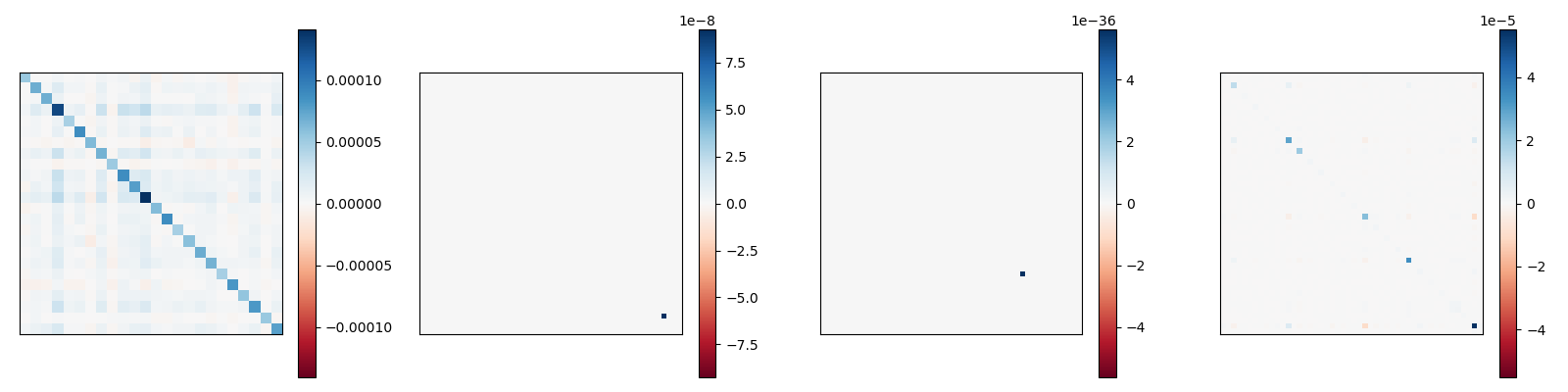}
    \includegraphics[width=0.9\textwidth]{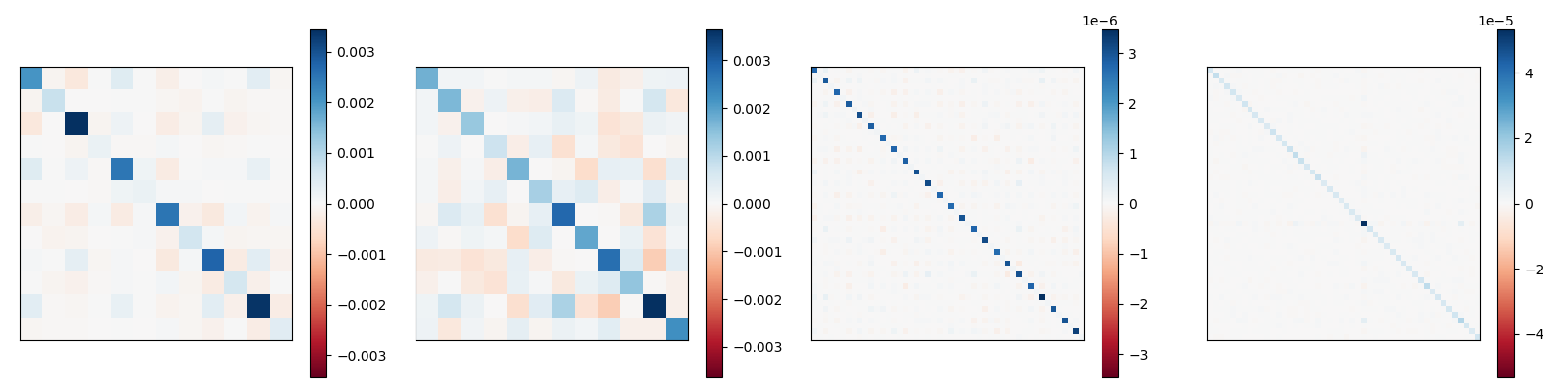}
    \caption{\textit{Raw layer-wise FIMs for glow models}. For each row of plots, we randomly select $4$ layers $\B{\theta}_\ell$ from glow models trained on (going from top to bottom) \texttt{SVHN}, \texttt{CelebA}, \texttt{GTSRB}, \texttt{CIFAR-10} and \texttt{ImageNet32}. We then plot the raw FIM $F_{\alpha \beta}$ values for $\max(50, \lvert \B{\theta} \rvert)$ weights in these layers, using a separate colorbar per layer to account for the fact that the absolute sizes of the FIM elements vary by orders of magnitudes from layer to layer.}
    \label{fig:raw FIM glow}
\end{figure}

\subsubsection{Raw, single-layer FIMs of diffusion models}

\begin{figure}[H]
    \centering
    \includegraphics[width=0.9\textwidth]{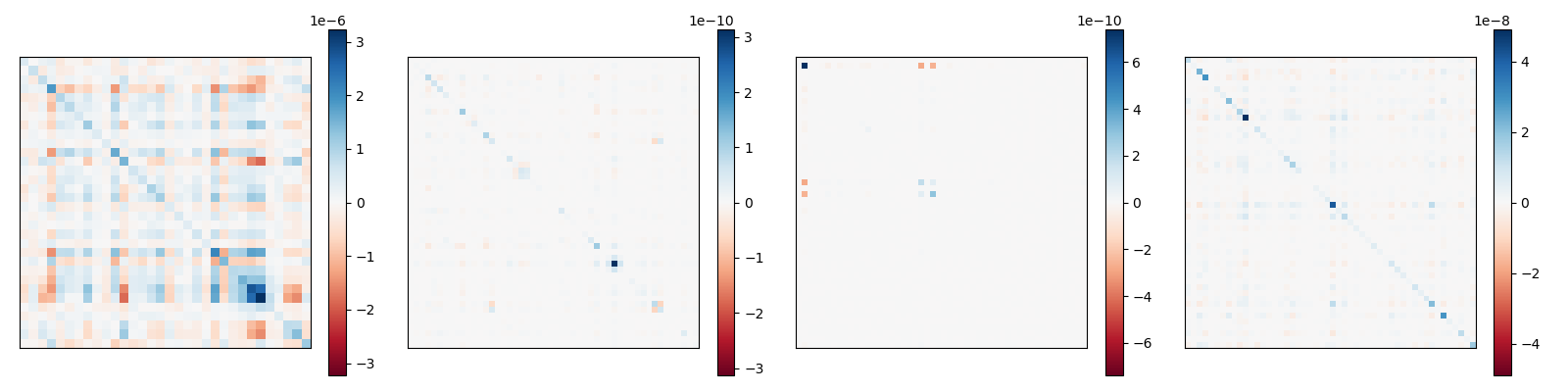}
    \includegraphics[width=0.9\textwidth]{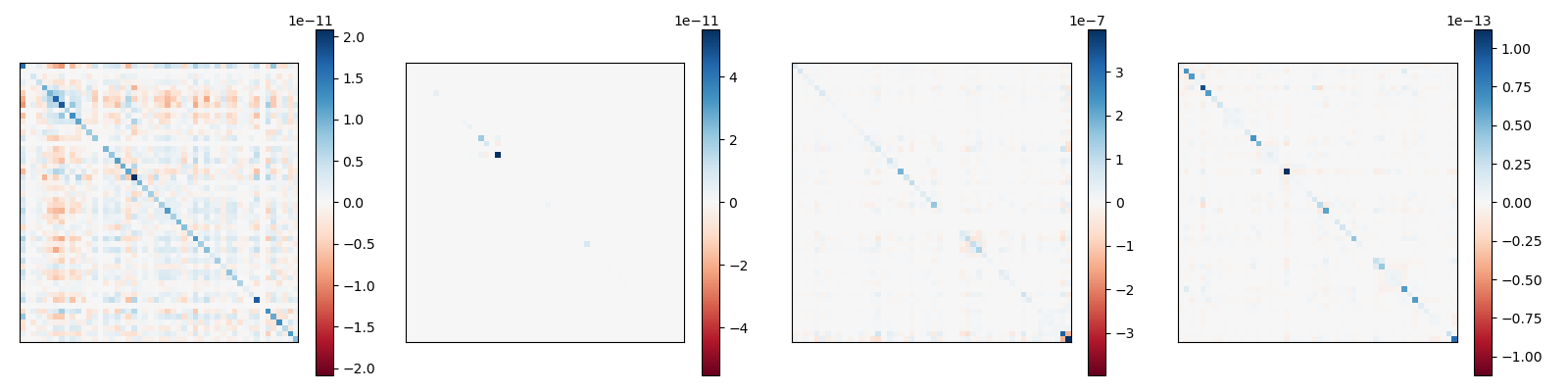} 
    \includegraphics[width=0.9\textwidth]{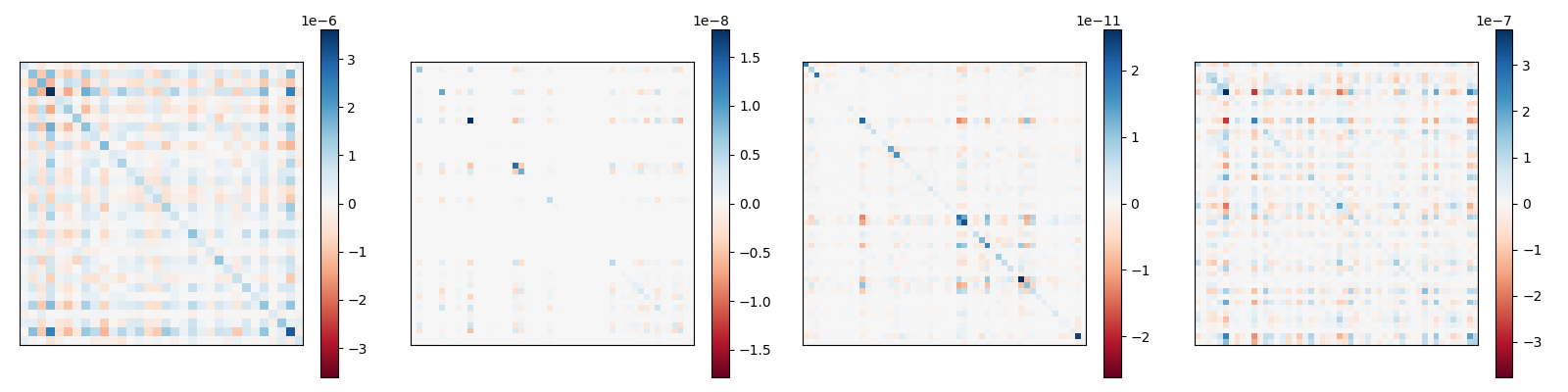}    \includegraphics[width=0.9\textwidth]{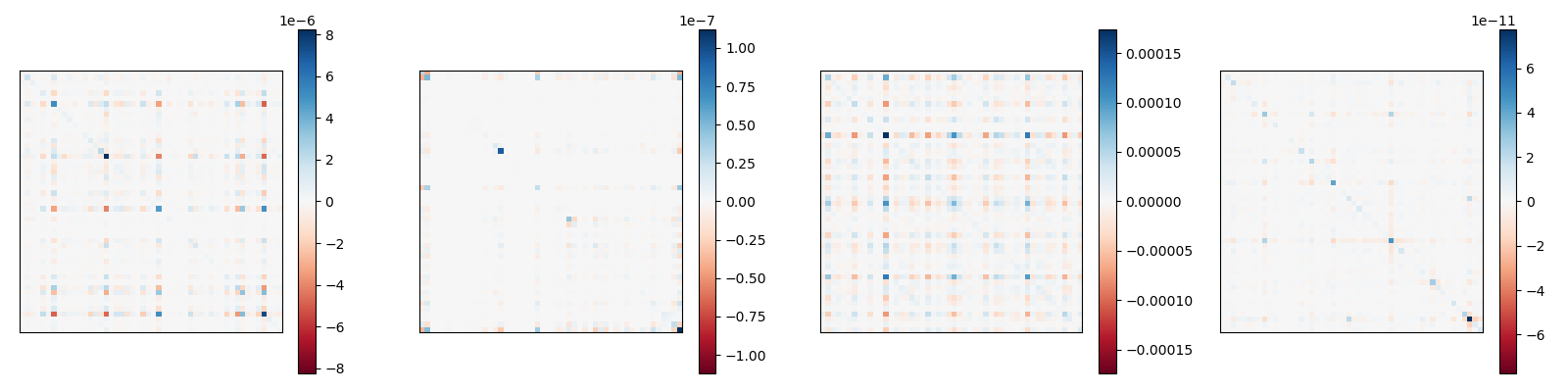}
    \includegraphics[width=0.9\textwidth]{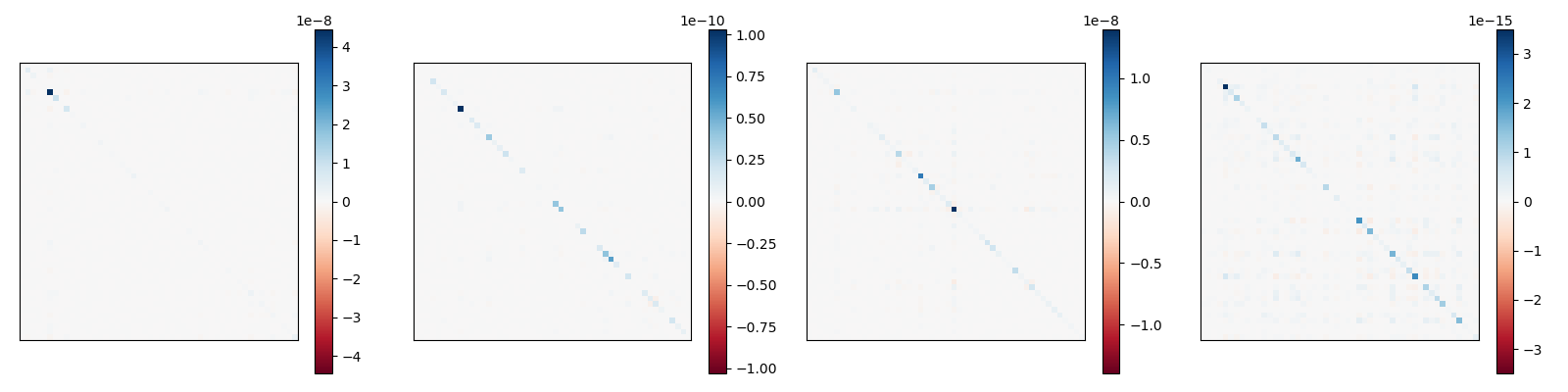}
    \caption{We replicate Figure \ref{fig:raw FIM glow} with a diffusion model (again using the gradient of the 1-timestep variational lower bound as a stand-in for the gradient of the log-likelihood), noting a qualitative difference in the appearance of the layers.}
    \label{fig:raw FIM diffusion}
\end{figure}

%% file: A7_diffusion_ablation_tables.tex
\subsection{Ablation study on likelihood proxies for diffusion models}
\label{app:diffusion ablation tables}

In this section, we discuss the use of different parts of a diffusion variational lower bound for anomaly detection. Before doing so, for completeness we present a condensed version of the theory presented in \cite{ho2020denoising}, using the same notation.

\subsubsection{Derivation of diffusion process}

Let our forward process be $q(\B{x}_t \vert \B{x_{t-1}})$, with prior of the true image distribution $q(\B{x}_0)$ and our learned reverse process be $p^{\B{\theta}}(\B{x}_{t-1} \vert \B{x}_{t})$ with Gaussian prior $p(\B{x}_T) \sim \mathcal{N}(\B{0}, \B{I})$. A variational lower bound on the model log-likelihood of can be derived \cite{ho2020denoising} as:

\begin{align}
    \mathbb{E}[- \log p^{\B{\theta}}(\B{x}_0)]
    \leq 
    \mathbb{E}[L_T + \sum_{t > 1} L_{t-1} + L_0] \\
    \text{where} \quad
    L_t = 
    \begin{cases}
        t = 0 
        & - \log p^{\B{\theta}}(\B{x}_0 \vert \B{x}_1) \\
        0 <  t < T 
        & KL(q(\B{x}_{t} \vert \B{x}_{t+1}, \B{x}_0) \vert \vert 
        p^{\B{\theta}}(\B{x}_{t} \vert \B{x}_{t+1})) \\
        t = T
        & KL(q(\B{x}_{T} \vert \B{x_0}) \vert \vert p(\B{x}_T))
    \end{cases}.
\end{align},

\cite{ho2020denoising} also derive that, if we parameterise our forward process as adding some normally distributed noise $\B{\epsilon}$  to $\B{x_{t-1}}$, and our reverse process as normally predicting this noise from $\B{x}_t$ via a network $\B{\epsilon}^{\B{\theta}}(\B{x_{t}}, t)$, for $0 \leq t < T$, $L_{t}$ can be computed as a squared error:

\begin{align}
    \mathbf{E} L_{t-1} = K_t \mathbf{E}_{\B{x}_t \sim q(\B{x}_0)} \norm{\B{\epsilon} - 
    \B{\epsilon}^{\B{\theta}}(\B{x_{t}}, t)}^2 + C_t.
    \label{eq:L_t definition}
\end{align}

Here $k_t$ and $C_t$ are constants independent of $\B{x}_0, \B{x}_1  \dots \B{x}_T$ and $\B{\theta}$, which can thus be omitted from computations. To compute the expectation, we use one sample from the reverse process $\B{x}_t \sim q(\B{x}_0)$, motivated by our findings in section \ref{app: multi q sample ablation} which show little to no performance gains from using five samples. Thus, we define our set of likelihood proxies as in (\ref{eq:L_t definition}) as $L_t(\B{x}) =  \norm{\B{\epsilon} - \B{\epsilon}^{\B{\theta}}(\B{x_{t+1}}, t+1)}$ for a single sample of noise $\B{\epsilon}$ from the reverse process $\B{x}_{t+1} \sim q(\B{x})$. Note that computing $L_t(\B{x})$ requires only one pass through the network, making it very efficient to compute. In section \ref{app: diffusion timestep ablation} we do an ablation study on the value of $t$ used, motivating our choice of $L_0$ in the application our method.

\subsubsection{On Representation Dependence in Diffusion models}

\begin{figure}[t]
    \centering
    \includegraphics[width=.48\textwidth]
{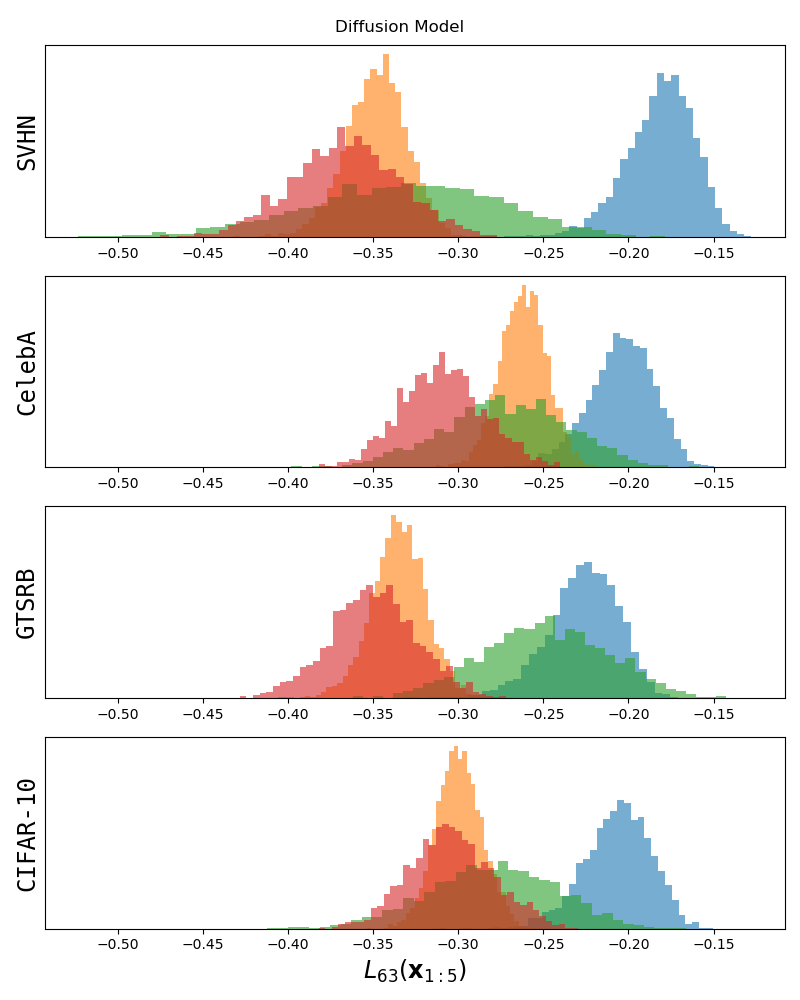}
    \includegraphics[width=.48\textwidth]
    {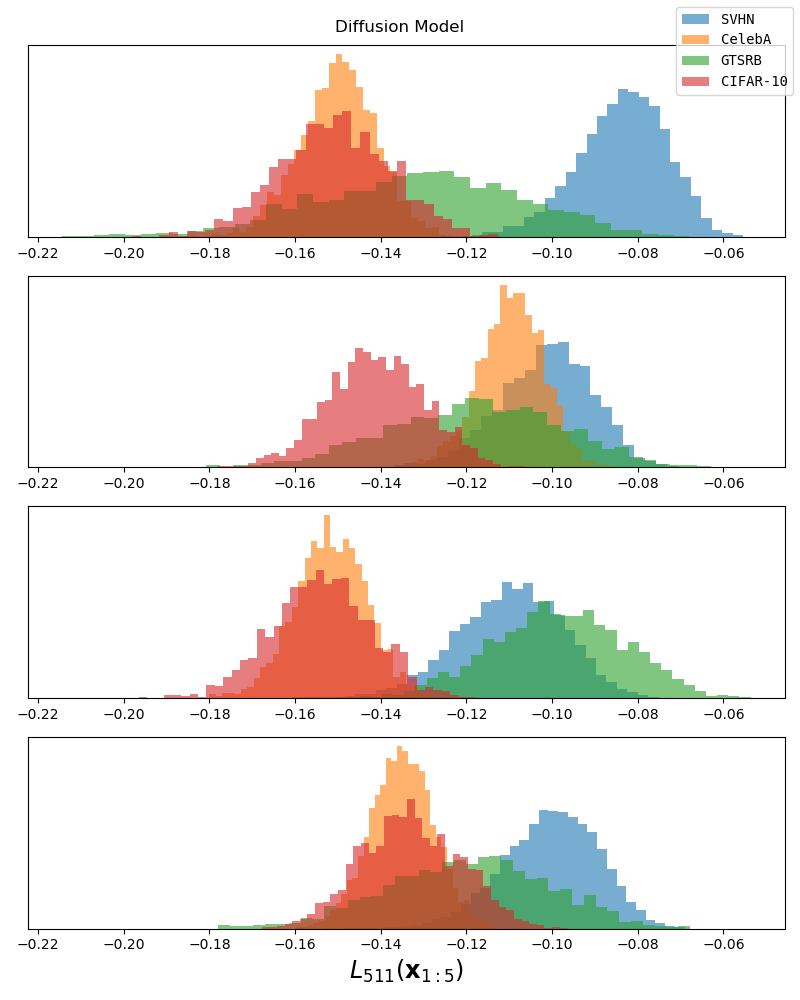}
    \caption{\emph{$L_{t-1}$ follows the pattern from \cite{nalisnick2018deep} for a variety of $t$ values.} We replicate figure \ref{likelihoodHistogram} for $t = 64$ [Left] and $t = 512$ [Right] \emph{using batch size $B=5$} (we use this batch size for reasons of limited compute.)
    }
    \label{fig:diffusion t likelihoodHistograms}
\end{figure}

When considering \cite{lan2021}'s results pertaining to representation dependence in the context of diffusion models, we arrive at the interesting question as to whether the choice of representation should be considered to affect the underlying distribution of the forward process $q$. Clearly the value we use in our method, $L_0 = \modeldist (\B{x}_0 \vert \B{x}_1)$, is representation-dependent. In the strict sense, the values $L_t$ for $t > 0$ aren't representation dependent, unless representation dependence is also considered to affect $q$, in which case this becomes more ambiguous. In figure \ref{fig:diffusion t likelihoodHistograms} we report the negative result that the values of $L_t$ for $t > 0$ also follow the pattern from \cite{nalisnick2018deep}, whereby structured OOD data has higher values for $L_t$. We defer further debate on this issue to future work.

\subsubsection{Ablation study on the value of $t$ used for anomaly detection with diffusion models}

\label{app: diffusion timestep ablation}

\begin{figure}
    \centering
    \includegraphics[width=0.5\textwidth]{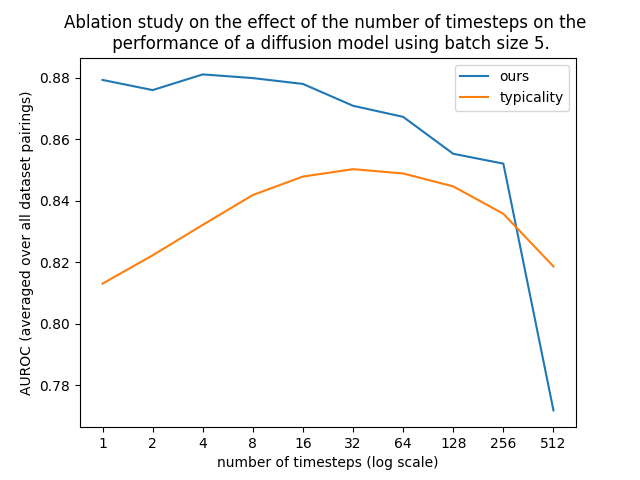}
    \caption{\textit{For diffusion models, $L_{t-1}$ is most informative for anomaly detection for low values of $t$.} We compute the AUROC values for all in/out dataset distribution pairings using $t = 2^n$ for $n = 0, 1 \dots 9$ and batch size $B = 5$ (for reasons of limited compute).}
    \label{fig:diffusion timestep ablation}
\end{figure}

In this section, we evaluate using different values of $t$ for the likelihood proxy $l(\B{x}) = L_{t-1}(\B{x})$ which we use as input for our anomaly detection method and typicality \cite{nalisnick2018deep}. We summarise our results in Figure \ref{fig:diffusion timestep ablation} by plotting the average AUROC acheived for each method across all $20$ dataset pairings. In table \ref{tab: diffusion timestep ablation} we provide more granular results with the AUROC for each pairing individually. We note the intuitive result that the performance of our method gradually decays as $t$ increases, corresponding to more noise being added to the sample fed into the network. Overall, the average performance of our method at $t = 1$ is higher than that of typicality which achieves its maximum performance at $t = 32$ (out of our model's maximum timestep of $T = 1000$). To ease compute requirements, we use batch size $B=5$ for all experiments.

\begin{table}[H]
\caption{auc values for typicality [top] and ours [bottom], batch size 5 applied to diffusion models for varied timesteps $t = 8, 64, 512$}
\begin{tabular}{l | l | r r r r r}
\toprule
 &  & \texttt{SVHN} & \texttt{CelebA} & \texttt{GTSRB} & \texttt{CIFAR-10} & \texttt{ImageNet32} \\
\midrule
\multirow[c]{5}{*}{\rotatebox{90}{\parbox{1.5cm}{typicality \\ $(t = 8)$ }}} & \texttt{SVHN} & - & 0.9937 & 0.7122 & \textbf{0.9975} & \textbf{0.9969} \\
 & \texttt{CelebA} & \textbf{1.0000} & - & 0.8816 & 0.3443 & \textbf{0.5904} \\
 & \texttt{GTSRB} & \textbf{0.9990} & 0.8108 & - & 0.6680 & 0.7214 \\
 & \texttt{CIFAR-10} & \textbf{1.0000} & 0.8684 & \textbf{0.9172} & - & 0.4852 \\
 & \texttt{ImageNet32} & \textbf{1.0000} & 0.9869 & \textbf{0.9845} & 0.8800 & - \\
\cline{1-7}
\multirow[c]{5}{*}{\rotatebox{90}{\parbox{1.5cm}{ours \\ $(t = 8)$ }}} & \texttt{SVHN} & - & \textbf{1.0000} & \textbf{0.9833} & 0.8214 & 0.9838 \\
 & \texttt{CelebA} & 0.9932 & - & \textbf{0.9284} & \textbf{0.8250} & 0.3942 \\
 & \texttt{GTSRB} & 0.9822 & \textbf{0.9998} & - & \textbf{0.7025} & \textbf{0.7485} \\
 & \texttt{CIFAR-10} & 0.9940 & \textbf{0.9998} & 0.8750 & - & \textbf{0.5071} \\
 & \texttt{ImageNet32} & 0.9934 & \textbf{1.0000} & 0.9624 & \textbf{0.9046} & - \\
\cline{1-7}

\midrule
\multirow[c]{5}{*}{\rotatebox{90}{\parbox{1.5cm}{typicality \\ $(t = 64)$ }}} & \texttt{SVHN} & - & 0.9813 & 0.4779 & \textbf{0.9970} & \textbf{0.9982} \\
 & \texttt{CelebA} & \textbf{1.0000} & - & \textbf{0.9746} & 0.3288 & \textbf{0.6622} \\
 & \texttt{GTSRB} & \textbf{0.9966} & 0.7805 & - & 0.6914 & \textbf{0.8543} \\
 & \texttt{CIFAR-10} & \textbf{1.0000} & 0.9228 & \textbf{0.9798} & - & \textbf{0.5388} \\
 & \texttt{ImageNet32} & \textbf{1.0000} & 0.9907 & \textbf{0.9965} & \textbf{0.8059} & - \\
\cline{1-7}
\multirow[c]{5}{*}{\rotatebox{90}{\parbox{1.5cm}{ours \\ $(t = 64)$ }}} & \texttt{SVHN} & - & \textbf{1.0000} & \textbf{0.9801} & 0.9345 & 0.9603 \\
 & \texttt{CelebA} & 0.9786 & - & 0.8856 & \textbf{0.6389} & 0.5063 \\
 & \texttt{GTSRB} & 0.9762 & \textbf{0.9990} & - & \textbf{0.7842} & 0.7150 \\
 & \texttt{CIFAR-10} & 0.9825 & \textbf{0.9996} & 0.7911 & - & 0.5039 \\
 & \texttt{ImageNet32} & 0.9850 & \textbf{0.9999} & 0.9255 & 0.8001 & - \\
\cline{1-7}

\midrule
\multirow[c]{5}{*}{\rotatebox{90}{\parbox{1.5cm}{typicality \\ $(t = 512)$ }}} & \texttt{SVHN} & - & 0.7125 & 0.5117 & \textbf{0.9574} & \textbf{0.9839} \\
 & \texttt{CelebA} & \textbf{0.9997} & - & \textbf{0.9984} & 0.3592 & 0.4543 \\
 & \texttt{GTSRB} & \textbf{0.9549} & 0.7854 & - & 0.7207 & \textbf{0.8194} \\
 & \texttt{CIFAR-10} & \textbf{0.9997} & 0.9815 & \textbf{0.9971} & - & \textbf{0.5154} \\
 & \texttt{ImageNet32} & \textbf{0.9998} & \textbf{0.9908} & \textbf{0.9984} & \textbf{0.6338} & - \\
\cline{1-7}
\multirow[c]{5}{*}{\rotatebox{90}{\parbox{1.5cm}{ours \\ $(t = 512)$ }}} & \texttt{SVHN} & - & \textbf{0.9962} & \textbf{0.9712} & 0.8089 & 0.7857 \\
 & \texttt{CelebA} & 0.7654 & - & 0.9396 & \textbf{0.5205} & \textbf{0.5272} \\
 & \texttt{GTSRB} & 0.6870 & \textbf{0.9496} & - & \textbf{0.7770} & 0.6256 \\
 & \texttt{CIFAR-10} & 0.6295 & \textbf{0.9844} & 0.9326 & - & 0.4660 \\
 & \texttt{ImageNet32} & 0.6635 & 0.9757 & 0.8904 & 0.5432 & - \\
\bottomrule
\end{tabular}
\label{tab: diffusion timestep ablation}
\end{table}

%% file: A3_background.tex
\section{Supervised gradient-based methodology for classifiers}

\label{sec: classifier gradient methodology}

For completeness, we discuss classifier based OOD detection methods using the gradient, noting that these methods \emph{are} given label-information at train time and our representation-invariance result does not directly translate over to this paradigm.

In order to compute gradients without a target label, 
This approach is hence supervised with respect to in-distribution and OOD labels which it requires.
\cite{liang2018enhancing} propose a method called ODIN which uses the gradient with respect to the \emph{data}: 
They backpropagate gradients to the input data to see how much an input perturbation can change the softmax output of a classifier, following the intuition that OOD inputs may be more sensitive and prone to a larger variation in the output distribution. 
\cite{igoe2022useful} are critical of the use of classifier gradients, instead advocating that most information can be recovered from the layer representations.
\cite{behpour2023gradorth} propose projection of the gradient onto the space generated by in-distribution gradients, motivated as in \cite{kwon2020backpropagated} by the informativity of the gradient angle for OOD detection.

%% file: A6_code.tex
\section{Code, models}
\label{app:Code, computational requirements, existing assets.}

\begin{figure}
    \centering
    \includegraphics[width=0.5\textwidth]{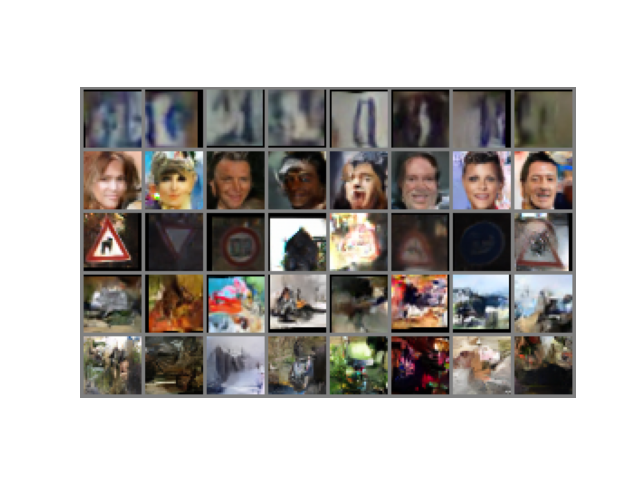}
    \caption{\textit{Samples from Glow models}. Samples from the Glow models used in our Experiments, trained on \texttt{SVHN}, \texttt{CelebA}, \texttt{GTSRB}, \texttt{CIFAR-10} and \texttt{ImageNet32} respectively from top to bottom.}
    \label{fig:samples from glow}
\end{figure}

Our \textit{Glow} implementation derives from a repository \url{https://github.com/y0ast/Glow-PyTorch} replicating the one used in \cite{nalisnick2018deep}, with the only difference being we use a batch size of 64 in training rather than 512. See Figure \ref{fig:samples from glow} for samples from our models.

\begin{figure}
    \centering
    \includegraphics[width=0.5\textwidth]{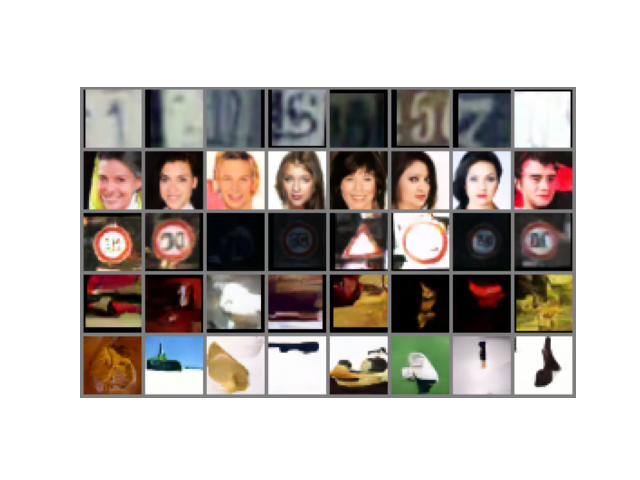}
    \caption{\textit{Samples from diffusion models}. Samples from the denoising diffusion models used in our Experiments, trained on \texttt{SVHN}, \texttt{CelebA}, \texttt{GTSRB}, \texttt{CIFAR-10} and \texttt{ImageNet32} respectively from top to bottom.}
    \label{fig:samples from diffusion}
\end{figure}

Our diffusion model implementation derives from a PyTorch transcription at \url{https://github.com/lucidrains/denoising-diffusion-pytorch} of that described in \cite{ho2020denoising}. We train using Adam with a learning rate of $3e^{-4}$ for $10$ epochs. Our model has $T=1000$ timesteps which are uniformly sampled from in training and the U-Net backbone has dimension multiplicities of $(1, 2, 4, 8)$.  See Fig \ref{fig:samples from diffusion} for samples from our models.

Our code is available with pre-trained models and pre-computed gradient $L^2$ norms at \githubRepo.

%% file: A11_VAE_results.tex
\newpage

\section{Results for poorly performant VAE}

\label{sec:VAEresults}

\begin{table}[H]
\caption{\textit{VAE models} Comparison of the AUROC values (larger values are better) of our method to the typicality test \cite{nalisnick2019} for batch sizes $B = 1, 5$. We train VAE \cite{kingma2014vae} models on five natural image datasets (as columns) and evaluate the ability of the model-method combination to reject the other datasets (as rows).}
\begin{tabular}{l | l | r r r r r}
\toprule
\multicolumn{2}{l |}{\textit{test} $\downarrow$  \textit{train} $\rightarrow$}
 & \texttt{SVHN} & \texttt{CelebA} & \texttt{GTSRB} & \texttt{CIFAR-10} & \texttt{ImageNet32} \\
\midrule
\multirow[c]{5}{*}{\rotatebox{90}{\parbox{1.5cm}{typicality \\ $(B = 1)$ }}}
 & \texttt{SVHN} & - & 0.5680 & 0.4302 & 0.5664 & \textbf{0.5933} \\
 & \texttt{CelebA} & \textbf{0.4569} & - & 0.3729 & \textbf{0.5085} & \textbf{0.4868} \\
 & \texttt{GTSRB} & 0.5901 & 0.6290 & - & 0.6484 & 0.6061 \\
 & \texttt{CIFAR-10} & 0.4223 & 0.4742 & 0.3454 & - & \textbf{0.4922} \\
 & \texttt{ImageNet32} & 0.4429 & 0.4831 & 0.3736 & 0.5047 & - \\
\cline{1-7}
\multirow[c]{5}{*}{\rotatebox{90}{\parbox{1.5cm}{ours \\ $(B = 1)$ }}} 
 & \texttt{SVHN} & - & \textbf{0.6693} & \textbf{0.5541} & \textbf{0.5904} & 0.5740 \\
 & \texttt{CelebA} & 0.4329 & - & \textbf{0.4337} & 0.4685 & 0.4588 \\
 & \texttt{GTSRB} & \textbf{0.5920} & \textbf{0.6581} & - & \textbf{0.6570} & \textbf{0.6629} \\
 & \texttt{CIFAR-10} & \textbf{0.4343} & \textbf{0.5826} & \textbf{0.5123} & - & 0.4864 \\
 & \texttt{ImageNet32} & \textbf{0.4582} & \textbf{0.5941} & \textbf{0.5048} & \textbf{0.5187} & - \\
\midrule
\midrule
\multirow[c]{5}{*}{\rotatebox{90}{\parbox{1.5cm}{typicality \\ $(B = 5)$ }}}
 & \texttt{SVHN} & - & 0.9978 & 0.7943 & \textbf{0.9975} & \textbf{0.9961} \\
 & \texttt{CelebA} & \textbf{1.0000} & - & 0.7642 & 0.3156 & 0.3621 \\
 & \texttt{GTSRB} & \textbf{0.9998} & 0.8336 & - & 0.6809 & 0.5765 \\
 & \texttt{CIFAR-10} & \textbf{1.0000} & 0.7808 & \textbf{0.8332} & - & 0.4488 \\
 & \texttt{ImageNet32} & \textbf{1.0000} & 0.9866 & \textbf{0.9675} & \textbf{0.9266} & - \\
\cline{1-7}
\multirow[c]{5}{*}{\rotatebox{90}{\parbox{1.5cm}{ours \\ $(B = 5)$ }}}
 & \texttt{SVHN} & - & \textbf{1.0000} & \textbf{0.9970} & 0.8457 & 0.9561 \\
 & \texttt{CelebA} & 0.9908 & - & \textbf{0.8552} & \textbf{0.7734} & \textbf{0.4202} \\
 & \texttt{GTSRB} & 0.9716 & \textbf{0.9997} & - & \textbf{0.7325} & \textbf{0.9007} \\
 & \texttt{CIFAR-10} & 0.9895 & \textbf{0.9992} & 0.8104 & - & \textbf{0.5733} \\
 & \texttt{ImageNet32} & 0.9862 & \textbf{1.0000} & 0.9309 & 0.8532 & - \\
\bottomrule
\end{tabular}
\label{tab:VAEresults}
\end{table}

\begin{figure}
    \centering
    \includegraphics[width=0.5\textwidth]{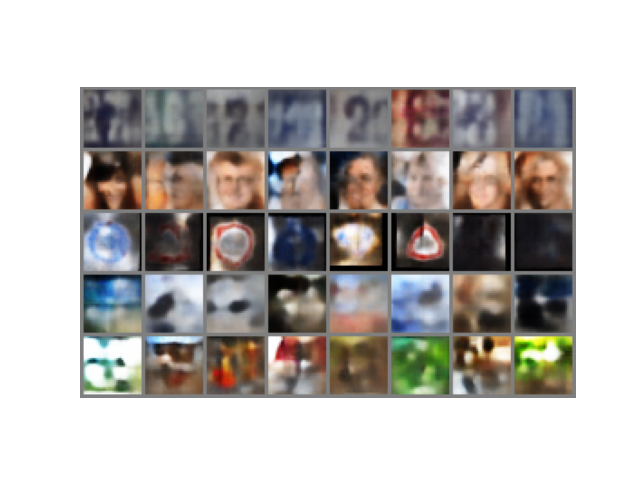}
    \caption{\textit{Samples from VAE models}. Samples from the denoising diffusion models used in our Experiments, trained on \texttt{SVHN}, \texttt{CelebA}, \texttt{GTSRB}, \texttt{CIFAR-10} and \texttt{ImageNet32} respectively from top to bottom.}
    \label{fig:samples from vae}
\end{figure}

Our VAE \cite{kingma2014vae} implementation uses entirely convolutional layers, in Figure \ref{fig:samples from vae} we note that the samples produced approximate the colour palate of the train datasets well but have poor semantic coherence. 
In table \ref{tab:VAEresults} we note poor performance for both our methods using this models as a backbone.

%% file: 0_main.bbl
\begin{thebibliography}{54}
\providecommand{\natexlab}[1]{#1}
\providecommand{\url}[1]{\texttt{#1}}
\expandafter\ifx\csname urlstyle\endcsname\relax
  \providecommand{\doi}[1]{doi: #1}\else
  \providecommand{\doi}{doi: \begingroup \urlstyle{rm}\Url}\fi

\bibitem[Amari(1998)]{amari1998naturalgrad}
Shun-ichi Amari.
\newblock Natural gradient works efficiently in learning.
\newblock \emph{Neural Computation}, 10\penalty0 (2):\penalty0 251--276, 1998.

\bibitem[Bartlett \& Kendall(1946)Bartlett and Kendall]{bartletts1946chisquare}
M.~S. Bartlett and D.~G. Kendall.
\newblock The statistical analysis of variance-heterogeneity and the logarithmic transformation.
\newblock \emph{Supplement to the Journal of the Royal Statistical Society}, 8\penalty0 (1):\penalty0 128--138, 1946.

\bibitem[Baur et~al.(2021)Baur, Denner, Wiestler, Navab, and Albarqouni]{baur2021autoencoders}
Christoph Baur, Stefan Denner, Benedikt Wiestler, Nassir Navab, and Shadi Albarqouni.
\newblock Autoencoders for unsupervised anomaly segmentation in brain mr images: a comparative study.
\newblock \emph{Medical Image Analysis}, 69:\penalty0 101952, 2021.

\bibitem[Behpour et~al.(2023)Behpour, Doan, Li, He, Gou, and Ren]{behpour2023gradorth}
Sima Behpour, Thang Doan, Xin Li, Wenbin He, Liang Gou, and Liu Ren.
\newblock Gradorth: A simple yet efficient out-of-distribution detection with orthogonal projection of gradients, arXiv, 2023.

\bibitem[Bergamin et~al.(2022)Bergamin, Mattei, Havtorn, Senetaire, Schmutz, Maal{\o}e, Hauberg, and Frellsen]{bergamin2022model}
Federico Bergamin, Pierre-Alexandre Mattei, Jakob~Drachmann Havtorn, Hugo Senetaire, Hugo Schmutz, Lars Maal{\o}e, Soren Hauberg, and Jes Frellsen.
\newblock Model-agnostic out-of-distribution detection using combined statistical tests.
\newblock In \emph{International Conference on Artificial Intelligence and Statistics}. PMLR, 2022.

\bibitem[Bishop(1994)]{bishop1994novelty}
Christopher~M Bishop.
\newblock Novelty detection and neural network validation.
\newblock \emph{IEE Proceedings-Vision, Image and Signal processing}, 141\penalty0 (4):\penalty0 217--222, 1994.

\bibitem[Blum et~al.(2020)Blum, Hopcroft, and Kannan]{gaussianAnnulus}
Avrim Blum, John Hopcroft, and Ravindran Kannan.
\newblock \emph{Foundations of Data Science}.
\newblock Cambridge University Press, 2020.

\bibitem[Caterini \& Loaiza-Ganem(2022)Caterini and Loaiza-Ganem]{caterini2022entropy}
Anthony~L. Caterini and Gabriel Loaiza-Ganem.
\newblock Entropic issues in likelihood-based {OOD} detection.
\newblock In Melanie~F. Pradier, Aaron Schein, Stephanie Hyland, Francisco J.~R. Ruiz, and Jessica~Z. Forde (eds.), \emph{Proceedings on "I (Still) Can't Believe It's Not Better!" at NeurIPS 2021 Workshops}. 13 Dec 2022.

\bibitem[Choi et~al.(2018)Choi, Jang, and Alemi]{choi2018waic}
Hyunsun Choi, Eric Jang, and Alexander~A Alemi.
\newblock Waic, but why? generative ensembles for robust anomaly detection.
\newblock \emph{arXiv preprint arXiv:1810.01392}, 2018.

\bibitem[Choi et~al.(2021)Choi, Yoon, Bae, and Kang]{choi2021robust}
Jaemoo Choi, Changyeon Yoon, Jeongwoo Bae, and Myungjoo Kang.
\newblock Robust out-of-distribution detection on deep probabilistic generative models, arXiv, 2021.

\bibitem[Cover \& Thomas(1991)Cover and Thomas]{typicalSetDefinition}
Thomas~M. Cover and Joy~A. Thomas.
\newblock \emph{Asymptotic Equipartition Property}, chapter~3, pp.\  57--69.
\newblock John Wiley \& Sons, Ltd, 1991.
\newblock ISBN 9780471748823.

\bibitem[Fisher(1938)]{Fisher_method_1938}
R.~A. Fisher.
\newblock \emph{Statistical methods for research workers}.
\newblock Oliver and Boyd, 1938.

\bibitem[Ganin et~al.(2016)Ganin, Ustinova, Ajakan, Germain, Larochelle, Laviolette, Marchand, and Lempitsky]{DANN}
Y.~Ganin, E.~Ustinova, H.~Ajakan, P.~Germain, H.~Larochelle, F.~Laviolette, M.~Marchand, and V.~Lempitsky.
\newblock Domain-adversarial training of neural networks.
\newblock \emph{Journal of Machine Learning Research}, 17:\penalty0 1–35, 2016.

\bibitem[Garg et~al.(2023)Garg, Erickson, Sharpnack, Smola, Balakrishnan, and Lipton]{garg2023rlsbench}
Saurabh Garg, Nick Erickson, James Sharpnack, Alex Smola, Sivaraman Balakrishnan, and Zachary~C Lipton.
\newblock Rlsbench: Domain adaptation under relaxed label shift.
\newblock \emph{arXiv preprint arXiv:2302.03020}, 2023.

\bibitem[George et~al.(2018)George, Laurent, Bouthillier, Ballas, and Vincent]{thomas2018ekfac}
Thomas George, C\'{e}sar Laurent, Xavier Bouthillier, Nicolas Ballas, and Pascal Vincent.
\newblock Fast approximate natural gradient descent in a kronecker factored eigenbasis.
\newblock In S.~Bengio, H.~Wallach, H.~Larochelle, K.~Grauman, N.~Cesa-Bianchi, and R.~Garnett (eds.), \emph{Advances in Neural Information Processing Systems}. 2018.

\bibitem[Graham et~al.(2023)Graham, Pinaya, Tudosiu, Nachev, Ourselin, and Cardoso]{Graham_2023_CVPR}
Mark~S. Graham, Walter~H.L. Pinaya, Petru-Daniel Tudosiu, Parashkev Nachev, Sebastien Ourselin, and Jorge Cardoso.
\newblock Denoising diffusion models for out-of-distribution detection.
\newblock In \emph{Proceedings of the IEEE/CVF Conference on Computer Vision and Pattern Recognition (CVPR) Workshops}, June 2023.

\bibitem[Havtorn et~al.(2021)Havtorn, Frellsen, Hauberg, and Maal{\o}e]{havtorn2021}
Jakob~D. Havtorn, Jes Frellsen, S{\o}ren Hauberg, and Lars Maal{\o}e.
\newblock Hierarchical vaes know what they don’t know.
\newblock In Marina Meila and Tong Zhang (eds.), \emph{Proceedings of the 38th International Conference on Machine Learning}. 18--24 Jul 2021.

\bibitem[Hendrycks \& Gimpel(2017)Hendrycks and Gimpel]{hendrycks2017a}
Dan Hendrycks and Kevin Gimpel.
\newblock A baseline for detecting misclassified and out-of-distribution examples in neural networks.
\newblock In \emph{International Conference on Learning Representations}, 2017.

\bibitem[Hendrycks et~al.(2018)Hendrycks, Mazeika, and Dietterich]{hendrycks2019}
Dan Hendrycks, Mantas Mazeika, and Thomas~G. Dietterich.
\newblock Deep anomaly detection with outlier exposure.
\newblock \emph{CoRR}, abs/1812.04606, 2018.

\bibitem[Ho et~al.(2020)Ho, Jain, and Abbeel]{ho2020denoising}
Jonathan Ho, Ajay Jain, and Pieter Abbeel.
\newblock Denoising diffusion probabilistic models.
\newblock \emph{Advances in Neural Information Processing Systems}, 33:\penalty0 6840--6851, 2020.

\bibitem[Igoe et~al.(2022)Igoe, Chung, Char, and Schneider]{igoe2022useful}
Conor Igoe, Youngseog Chung, Ian Char, and Jeff Schneider.
\newblock How useful are gradients for ood detection really?, arXiv, 2022.

\bibitem[Jaakkola \& Haussler(1998)Jaakkola and Haussler]{jaakkola1998exploiting}
Tommi Jaakkola and David Haussler.
\newblock Exploiting generative models in discriminative classifiers.
\newblock In M.~Kearns, S.~Solla, and D.~Cohn (eds.), \emph{Advances in Neural Information Processing Systems}. 1998.

\bibitem[Jacot et~al.(2018)Jacot, Gabriel, and Hongler]{jacot2018neural}
Arthur Jacot, Franck Gabriel, and Clement Hongler.
\newblock Neural tangent kernel: Convergence and generalization in neural networks.
\newblock In S.~Bengio, H.~Wallach, H.~Larochelle, K.~Grauman, N.~Cesa-Bianchi, and R.~Garnett (eds.), \emph{Advances in Neural Information Processing Systems}. 2018.

\bibitem[Kingma \& Ba(2015)Kingma and Ba]{kingma2015adam}
Diederik~P. Kingma and Jimmy Ba.
\newblock Adam: A method for stochastic optimization.
\newblock In \emph{International Conference on Learning Representations}, 2015.

\bibitem[Kingma \& Welling(2014)Kingma and Welling]{kingma2014vae}
Diederik~P Kingma and Max Welling.
\newblock Auto-encoding variational bayes.
\newblock In \emph{Proceedings of The 33rd International Conference on Machine Learning}, 2014.

\bibitem[Kingma \& Dhariwal(2018)Kingma and Dhariwal]{kingma2018glow}
Durk~P Kingma and Prafulla Dhariwal.
\newblock Glow: Generative flow with invertible 1x1 convolutions.
\newblock In S.~Bengio, H.~Wallach, H.~Larochelle, K.~Grauman, N.~Cesa-Bianchi, and R.~Garnett (eds.), \emph{Advances in Neural Information Processing Systems}. 2018.

\bibitem[Kirichenko et~al.(2020)Kirichenko, Izmailov, and Wilson]{kirichenko2020flows}
Polina Kirichenko, Pavel Izmailov, and Andrew~G Wilson.
\newblock Why normalizing flows fail to detect out-of-distribution data.
\newblock In H.~Larochelle, M.~Ranzato, R.~Hadsell, M.F. Balcan, and H.~Lin (eds.), \emph{Advances in Neural Information Processing Systems}. 2020.

\bibitem[Kwon et~al.(2020)Kwon, Prabhushankar, Temel, and AlRegib]{kwon2020backpropagated}
Gukyeong Kwon, Mohit Prabhushankar, Dogancan Temel, and Ghassan AlRegib.
\newblock Backpropagated gradient representations for anomaly detection.
\newblock In \emph{Proceedings of the European Conference on Computer Vision (ECCV)}, 2020.

\bibitem[Le~Lan \& Dinh(2021)Le~Lan and Dinh]{lan2021}
Charline Le~Lan and Laurent Dinh.
\newblock Perfect density models cannot guarantee anomaly detection.
\newblock \emph{Entropy}, 23\penalty0 (12), 2021.

\bibitem[Liang et~al.(2018)Liang, Li, and Srikant]{liang2018enhancing}
Shiyu Liang, Yixuan Li, and R.~Srikant.
\newblock Enhancing the reliability of out-of-distribution image detection in neural networks.
\newblock In \emph{International Conference on Learning Representations}, 2018.

\bibitem[Liu et~al.(2020)Liu, Wang, Owens, and Li]{liu2020energy}
Weitang Liu, Xiaoyun Wang, John Owens, and Yixuan Li.
\newblock Energy-based out-of-distribution detection.
\newblock \emph{Advances in neural information processing systems}, 33:\penalty0 21464--21475, 2020.

\bibitem[Martens(2020)]{martens2020naturalgrad}
James Martens.
\newblock New insights and perspectives on the natural gradient method.
\newblock \emph{Journal of Machine Learning Research}, 21\penalty0 (146):\penalty0 1--76, 2020.

\bibitem[Morningstar et~al.(2021)Morningstar, Ham, Gallagher, Lakshminarayanan, Alemi, and Dillon]{pmlr-v130-morningstar21a}
Warren Morningstar, Cusuh Ham, Andrew Gallagher, Balaji Lakshminarayanan, Alex Alemi, and Joshua Dillon.
\newblock Density of states estimation for out of distribution detection.
\newblock In Arindam Banerjee and Kenji Fukumizu (eds.), \emph{Proceedings of The 24th International Conference on Artificial Intelligence and Statistics}. 13--15 Apr 2021.

\bibitem[Nalisnick et~al.(2019{\natexlab{a}})Nalisnick, Matsukawa, Teh, Gorur, and Lakshminarayanan]{nalisnick2018deep}
Eric Nalisnick, Akihiro Matsukawa, Yee~Whye Teh, Dilan Gorur, and Balaji Lakshminarayanan.
\newblock Do deep generative models know what they don't know?
\newblock In \emph{International Conference on Learning Representations}, 2019{\natexlab{a}}.

\bibitem[Nalisnick et~al.(2019{\natexlab{b}})Nalisnick, Matsukawa, Teh, and Lakshminarayanan]{nalisnick2019}
Eric Nalisnick, Akihiro Matsukawa, Yee~Whye Teh, and Balaji Lakshminarayanan.
\newblock Detecting out-of-distribution inputs to deep generative models using typicality, arXiv, 2019{\natexlab{b}}.

\bibitem[Nguyen et~al.(2015)Nguyen, Yosinski, and Clune]{nguyen2015deep}
Anh Nguyen, Jason Yosinski, and Jeff Clune.
\newblock Deep neural networks are easily fooled: High confidence predictions for unrecognizable images.
\newblock In \emph{Proceedings of the IEEE conference on computer vision and pattern recognition}, 2015.

\bibitem[Nguyen et~al.(2019)Nguyen, Lim, Divakaran, Low, and Chan]{nguyen2019gee}
Quoc~Phong Nguyen, Kar~Wai Lim, Dinil~Mon Divakaran, Kian~Hsiang Low, and Mun~Choon Chan.
\newblock Gee: A gradient-based explainable variational autoencoder for network anomaly detection.
\newblock In \emph{2019 IEEE Conference on Communications and Network Security (CNS)}. IEEE, 2019.

\bibitem[Papamakarios et~al.(2021)Papamakarios, Nalisnick, Rezende, Mohamed, and Lakshminarayanan]{papamakarios2021normalizing}
George Papamakarios, Eric Nalisnick, Danilo~Jimenez Rezende, Shakir Mohamed, and Balaji Lakshminarayanan.
\newblock Normalizing flows for probabilistic modeling and inference.
\newblock \emph{The Journal of Machine Learning Research}, 22\penalty0 (1):\penalty0 2617--2680, 2021.

\bibitem[Radhakrishna~Rao(1948)]{rao_1948}
C.~Radhakrishna~Rao.
\newblock Large sample tests of statistical hypotheses concerning several parameters with applications to problems of estimation.
\newblock \emph{Mathematical Proceedings of the Cambridge Philosophical Society}, 44\penalty0 (1):\penalty0 50–57, 1948.

\bibitem[Ren et~al.(2019)Ren, Liu, Fertig, Snoek, Poplin, Depristo, Dillon, and Lakshminarayanan]{ren2019}
Jie Ren, Peter~J. Liu, Emily Fertig, Jasper Snoek, Ryan Poplin, Mark Depristo, Joshua Dillon, and Balaji Lakshminarayanan.
\newblock Likelihood ratios for out-of-distribution detection.
\newblock In H.~Wallach, H.~Larochelle, A.~Beygelzimer, F.~d\textquotesingle Alch\'{e}-Buc, E.~Fox, and R.~Garnett (eds.), \emph{Advances in Neural Information Processing Systems}. 2019.

\bibitem[Salimans et~al.(2017)Salimans, Karpathy, Chen, and Kingma]{salimans2017pixelcnn++}
Tim Salimans, Andrej Karpathy, Xi~Chen, and Diederik~P Kingma.
\newblock Pixelcnn++: Improving the pixelcnn with discretized logistic mixture likelihood and other modifications.
\newblock \emph{arXiv preprint arXiv:1701.05517}, 2017.

\bibitem[Schirrmeister et~al.(2020)Schirrmeister, Zhou, Ball, and Zhang]{Schirrmeister2020heirachy}
Robin Schirrmeister, Yuxuan Zhou, Tonio Ball, and Dan Zhang.
\newblock Understanding anomaly detection with deep invertible networks through hierarchies of distributions and features.
\newblock In H.~Larochelle, M.~Ranzato, R.~Hadsell, M.F. Balcan, and H.~Lin (eds.), \emph{Advances in Neural Information Processing Systems}. 2020.

\bibitem[Serrà et~al.(2020)Serrà, Álvarez, Gómez, Slizovskaia, Núñez, and Luque]{serra2019input}
Joan Serrà, David Álvarez, Vicenç Gómez, Olga Slizovskaia, José~F. Núñez, and Jordi Luque.
\newblock Input complexity and out-of-distribution detection with likelihood-based generative models.
\newblock In \emph{International Conference on Learning Representations}, 2020.

\bibitem[Sohl-Dickstein et~al.(2015{\natexlab{a}})Sohl-Dickstein, Weiss, Maheswaranathan, and Ganguli]{pmlr-v37-sohl-dickstein15}
Jascha Sohl-Dickstein, Eric Weiss, Niru Maheswaranathan, and Surya Ganguli.
\newblock Deep unsupervised learning using nonequilibrium thermodynamics.
\newblock In Francis Bach and David Blei (eds.), \emph{Proceedings of the 32nd International Conference on Machine Learning}, Lille, France, 07--09 Jul 2015{\natexlab{a}}. PMLR.

\bibitem[Sohl-Dickstein et~al.(2015{\natexlab{b}})Sohl-Dickstein, Weiss, Maheswaranathan, and Ganguli]{sohl2015deep}
Jascha Sohl-Dickstein, Eric Weiss, Niru Maheswaranathan, and Surya Ganguli.
\newblock Deep unsupervised learning using nonequilibrium thermodynamics.
\newblock In \emph{International conference on machine learning}. PMLR, 2015{\natexlab{b}}.

\bibitem[Stilgoe(2020)]{stilgoe2020uber}
Jack Stilgoe.
\newblock \emph{Who Killed Elaine Herzberg?}, pp.\  1--6.
\newblock Springer International Publishing, Cham, 2020.
\newblock ISBN 978-3-030-32320-2.

\bibitem[Szegedy et~al.(2014)Szegedy, Zaremba, Sutskever, Bruna, Erhan, Goodfellow, and Fergus]{szegedy2014intriguing}
Christian Szegedy, Wojciech Zaremba, Ilya Sutskever, Joan Bruna, Dumitru Erhan, Ian Goodfellow, and Rob Fergus.
\newblock Intriguing properties of neural networks.
\newblock January 2014.
\newblock 2nd International Conference on Learning Representations, ICLR 2014 ; Conference date: 14-04-2014 Through 16-04-2014.

\bibitem[Tieleman et~al.(2012)Tieleman, Hinton, et~al.]{tieleman2012rmsprop}
Tijmen Tieleman, Geoffrey Hinton, et~al.
\newblock Lecture 6.5-rmsprop: Divide the gradient by a running average of its recent magnitude.
\newblock \emph{COURSERA: Neural networks for machine learning}, 4\penalty0 (2):\penalty0 26--31, 2012.

\bibitem[Ulmer et~al.(2020)Ulmer, Meijerink, and Cin\`a]{ulmer2020trust}
Dennis Ulmer, Lotta Meijerink, and Giovanni Cin\`a.
\newblock Trust issues: Uncertainty estimation does not enable reliable ood detection on medical tabular data.
\newblock In Emily Alsentzer, Matthew B.~A. McDermott, Fabian Falck, Suproteem~K. Sarkar, Subhrajit Roy, and Stephanie~L. Hyland (eds.), \emph{Proceedings of the Machine Learning for Health NeurIPS Workshop}. 11 Dec 2020.

\bibitem[Van~den Oord et~al.(2016)Van~den Oord, Kalchbrenner, Espeholt, Vinyals, Graves, et~al.]{van2016conditional}
Aaron Van~den Oord, Nal Kalchbrenner, Lasse Espeholt, Oriol Vinyals, Alex Graves, et~al.
\newblock Conditional image generation with pixelcnn decoders.
\newblock \emph{Advances in neural information processing systems}, 29, 2016.

\bibitem[Xiao et~al.(2020)Xiao, Yan, and Amit]{zhisheng2020regret}
Zhisheng Xiao, Qing Yan, and Yali Amit.
\newblock Likelihood regret: An out-of-distribution detection score for variational auto-encoder.
\newblock In H.~Larochelle, M.~Ranzato, R.~Hadsell, M.F. Balcan, and H.~Lin (eds.), \emph{Advances in Neural Information Processing Systems}. 2020.

\bibitem[Zhang et~al.(2021{\natexlab{a}})Zhang, Goldstein, and Ranganath]{zhang2021understanding}
Lily Zhang, Mark Goldstein, and Rajesh Ranganath.
\newblock Understanding failures in out-of-distribution detection with deep generative models.
\newblock In \emph{International Conference on Machine Learning}. PMLR, 2021{\natexlab{a}}.

\bibitem[Zhang et~al.(2021{\natexlab{b}})Zhang, Zhang, and McDonagh]{zhang2021nelloc}
Mingtian Zhang, Andi Zhang, and Steven McDonagh.
\newblock On the out-of-distribution generalization of probabilistic image modelling.
\newblock In M.~Ranzato, A.~Beygelzimer, Y.~Dauphin, P.S. Liang, and J.~Wortman Vaughan (eds.), \emph{Advances in Neural Information Processing Systems}. 2021{\natexlab{b}}.

\bibitem[Zhou et~al.(2021)Zhou, Liu, Qiao, Xiang, and Loy]{zhou2021domain}
Kaiyang Zhou, Ziwei Liu, Yu~Qiao, Tao Xiang, and Chen~Change Loy.
\newblock Domain generalization in vision: A survey.
\newblock \emph{arXiv preprint arXiv:2103.02503}, 2021.

\end{thebibliography}
